\documentclass[runningheads]{llncs}

\usepackage{eccv}

\usepackage{eccvabbrv}

\usepackage{eccvabbrv}
\usepackage{graphicx}
\usepackage{booktabs}
\usepackage{multirow}
\usepackage{amsfonts}
\usepackage[accsupp]{axessibility}  
\usepackage{orcidlink}
\usepackage{algorithm}
\usepackage{algorithmicx}
\usepackage[noend]{algcompatible}
\usepackage{algpseudocode}
\usepackage{amsmath}
\usepackage{amssymb}
\usepackage[most]{tcolorbox}
\usepackage[table]{xcolor}
\usepackage{multirow, makecell, colortbl, xcolor}
\usepackage{xparse}
\usepackage[accsupp]{axessibility} 
\usepackage{enumitem}
\usepackage{tikz}          
\usepackage{tcolorbox}

\newlist{phaselist}{enumerate}{1}
\setlist[phaselist,1]{label=\textbf{Phase \arabic*:}, leftmargin=*, labelsep=0.6em, font=\large\bfseries, itemsep=1.5em}

\newlist{modulelist}{itemize}{1}
\setlist[modulelist,1]{label=\textbullet, leftmargin=0.01em, labelsep=0.5em, itemsep=0.8em}

\newlist{iolist}{itemize}{1}
\setlist[iolist,1]{label=--, leftmargin=1.0em, labelsep=0.5em, itemsep=0.3em}

\newcommand{\module}[1]{\textbf{Module: #1}}
\newcommand{\inp}[1]{\textit{Input:} #1}
\newcommand{\proc}[1]{\textit{Process:} #1}
\newcommand{\outp}[1]{\textit{Output:} #1}

\tcbuselibrary{breakable}

\newtcolorbox{codeout}{
  colback=gray!5, colframe=gray!35, boxrule=0.4pt, arc=2pt,
  left=6pt, right=6pt, top=4pt, bottom=4pt,
  breakable, fontupper=\ttfamily\small
}

\NewDocumentCommand{\pydict}{m}{%
  \begin{codeout}
  \begin{itemize}[leftmargin=1em, label={}, itemsep=2pt, topsep=2pt, parsep=0pt]
  \foreach \k/\v in {#1}{\item \textbf{\k}: \v}
  \end{itemize}
  \end{codeout}
}

\NewDocumentCommand{\pylist}{m}{%
  \begin{codeout}
  [\foreach \x [count=\i] in {#1}{\ifnum\i>1, \fi\x}]
  \end{codeout}
}

\tcbuselibrary{listings, skins, breakable}

\newtcblisting{codelist}{
  colback=gray!5, colframe=gray!35, boxrule=0.4pt, arc=2pt,
  left=6pt, right=6pt, top=4pt, bottom=4pt,
  breakable, listing only,
  listing options={
    basicstyle=\ttfamily\small,
    breaklines=true,
    breakatwhitespace=false,
    columns=fullflexible,
    showstringspaces=false,
    tabsize=2
  }
}

\usepackage{hyperref}

\usepackage{orcidlink}
\newtcolorbox{radiologistcomment}{
colback=gray!10,
colframe=gray!60,
coltitle=black,
boxrule=0.5pt,
arc=2mm,
left=6pt,
right=6pt,
top=6pt,
bottom=6pt,
title=Radiologist Comment
}
\newtcolorbox{promptbox}[1][]{
  colback=gray!5,    
  colframe=gray!30,  
  coltitle=black,    
  fonttitle=\bfseries,
  title={#1},
  sharp corners,     
  boxrule=0.5pt,
  fontupper=\small\ttfamily, 
  breakable          
  }

\begin{document}

\title{MedPAO-VL: Interpretable and Protocol-Compliant Agent for Structured Radiology Report Generation} 

\titlerunning{MedPAO-VL}

\author{Shrish Shrinath Vaidya\inst{1}\orcidlink{0009-0006-5438-9500}\thanks{Corresponding author: \email{shrish.s.vaidya@dsai.iitm.ac.in}} \and
        Gowthamaan Palani\inst{1}\orcidlink{0009-0009-4095-2413} \and
        Minmini Selvam\inst{2} \and 
        Gokulraja Srinivasaraja\inst{3} \and 
        Ganapathy Krishnamurthi\inst{1}
        }

\authorrunning{S. S. Vaidya et al.}

\institute{
Department of Data Science and Artificial Intelligence,
Indian Institute of Technology Madras, Chennai, India\\
\and
TANUH -- The AI Centre of Excellence in Healthcare,
Indian Institute of Science, Bengaluru, India\\
\and
Department of Radiology (Division of Interventional Radiology), 
Ramaiah Medical College and Memorial hospital, Bengaluru, India\\
}

\maketitle
\begin{abstract}
Despite the remarkable generalization capabilities of Large Language Models, generating clinically accurate and structured radiology reports from medical images remains a significant challenge. Current end-to-end approaches function as black boxes, prone to factual hallucinations and generating ungrounded findings. This lack of reliability and failure to adhere to domain-specific structuring protocols, fundamentally limits their clinical utility. We present MedPAO-VL, a multimodal agent that shifts the paradigm from monolithic generation to explainable, iterative, protocol-compliant structured report generation. Unlike deterministic models, our plan-act-observe loop functions as a goal-oriented agent: it first perceives the Chest X-Ray through an explainable vision module designed to capture inherent diagnostic uncertainty, then iteratively refines the generated report by observing semantic discrepancies and self-correcting its generation for better alignment with visually detected clinical findings. Evaluation of our framework demonstrates that this agentic reasoning improves clinical correctness and completeness, achieving an expert-rated accuracy score of 3.5/5 and a structural score of 4.3/5. By prioritizing pathology specific explainability and iterative refinement over opaque end-to-end mapping, MedPAO-VL offers a transparent and clinically viable alternative for automated structured radiology report generation. The code is available at: \url{https://github.com/MiRL-IITM/medpao-vl}

\keywords{Explainability \and Structured Report Generation \and Multimodal agent \and Clinical Protocols \and Iterative refinement }

\end{abstract}
\section{Introduction}
\label{section:introduction}
\begin{figure}[!t]
    \centering
    \includegraphics[trim=0cm 0cm 0cm 3.8cm, clip,width=0.9\linewidth]{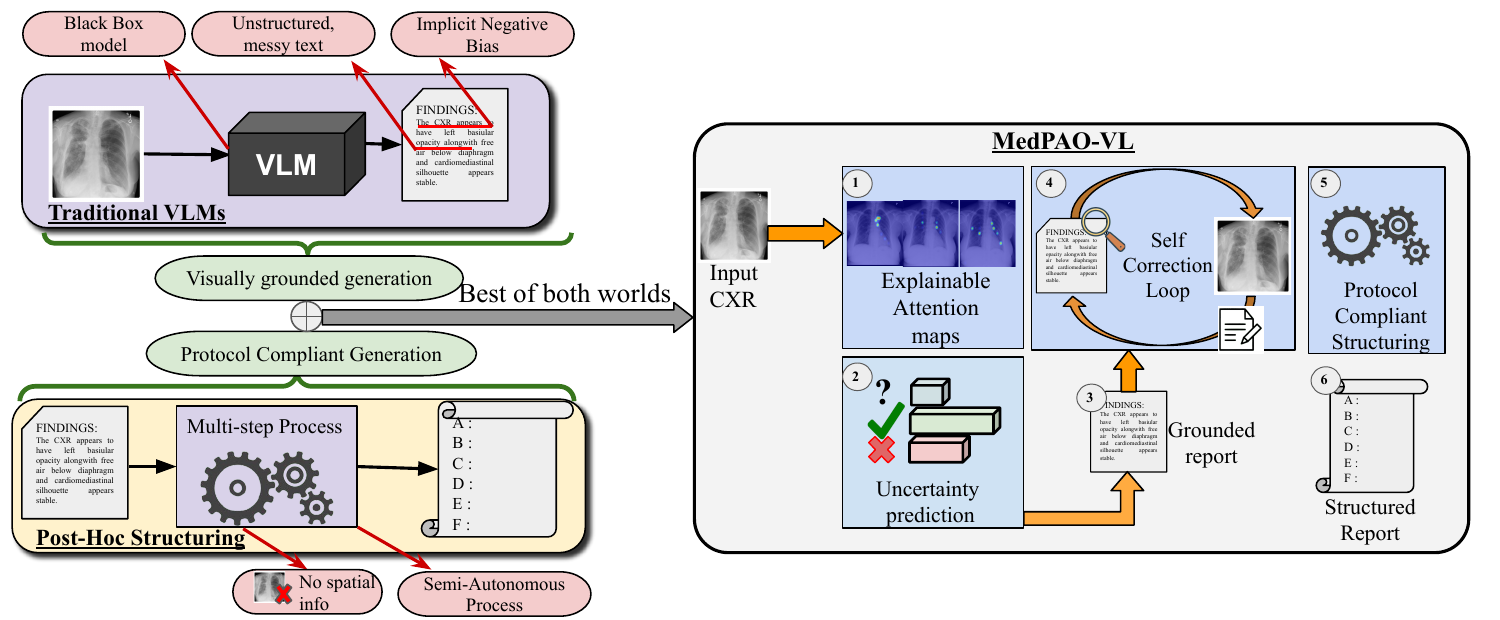}
    \caption{MedPAO-VL integrates explainable, uncertainty-aware visual perception and iterative self-correction, overcoming limitations of traditional VLMs and post-hoc report structuring methods.}
    \label{fig:fig1}
\end{figure}

Generating accurate and structured radiology reports remains a significant bottleneck in clinical workflows. Escalating volumes of medical imaging procedures exert immense pressure on radiologists, creating a critical need for automated systems that can reduce turnaround times and standardize reporting quality\cite{hong2025radiologist}. Yet in high-stakes clinical settings, automation alone is insufficient. Any deployed system must also be \textit{trustworthy}: its outputs must be \textit{interpretable}, its reasoning \textit{auditable}, and its failure modes clinically safe. This demand has catalyzed two parallel, yet largely distinct, streams of research: end-to-end report generation from images and the post-hoc structuring of unstructured reports. The first stream is dominated by multimodal models that generate reports directly from medical images. While these models demonstrate considerable promise, they function as black-boxes, offering clinicians little insight into how a conclusion was reached. A further fundamental limitation lies in their training data, which operates on the flawed assumption that any pathology not explicitly mentioned in a ground-truth report is implicitly absent. This is clinically unsound: the absence of a finding may reflect a lack of observation rather than its confirmed absence, and encoding this ambiguity as ground truth undermines both model reliability and clinical trust. In contrast, the second stream prioritizes verifiable reasoning and strict protocol compliance. Frameworks such as MedPAO\cite{medpao} employ an agentic loop to parse unstructured reports and transform them into a structured, protocol-compliant format like ABCDEF\cite{abcdef}(details in Supplementary material), producing outputs that are auditable and adherent to standardized guidelines. However, its primary limitation is its post-hoc nature: it structures existing text rather than reasoning from primary visual evidence, leaving the opacity of the upstream generative process unaddressed.

This dichotomy exposes a critical gap at the intersection of explainability and generative capability. No existing solution integrates the visual reasoning power of multimodal models with the transparent, protocol-driven auditability of agentic frameworks, a gap that directly compromises the trustworthiness of automated radiology pipelines. To bridge this gap, we introduce \textbf{MedPAO-VL} (MedPAO-Vision-Language), a multimodal agent which is both end-to-end and transparent by design to generate structured, protocol-compliant reports from CXRs. By extending the MedPAO framework\cite{medpao} with multimodal capabilities, we embed explainability into the generative process itself rather than as an afterthought. The key contributions of this work are as follows:

\begin{enumerate}
    \item We introduce \textbf{MedPAO-VL}, a multimodal agent that unifies visual perception with a Plan-Act-Observe loop, yielding protocol-compliant report generation with step-wise auditable reasoning.
    \item A novel \textbf{Vision Concept Module} employing a uncertainty aware labeling scheme for pathologies to provide reliable and explainable visual recognition that more faithfully reflects clinical epistemology.
    \item Extensive evaluation demonstrating that our hybrid architecture produces clinically sound reports that are factually consistent and protocol-compliant alongwith interpretable vision module, advancing the trustworthiness of automated radiology reporting.
\end{enumerate}
\section{Related Works}
\label{section:related_works}

\subsection{Radiology Report Generation}

Automated radiology report generation has emerged as a significant area of research, drawing initial inspiration from the successes of natural image captioning\cite{show_attend_tell, att2in, img_attn3}. Foundational approaches to report generation established the encoder-decoder framework, initially employing Convolutional Neural Networks (CNNs) paired with Long Short-Term Memory (LSTM) networks\cite{mti_cnn1, sdc_cnn2, hrg_cnn3}. With the advent of the Transformer\cite{Vaswani2017AttentionIA} architecture, the field evolved to replace recurrent decoders with LLMs, yielding CNN-LLM configurations\cite{sgf_cnn4, kuit_cnn5, m2kt_cnn6}. This evolution has culminated in Vision Transformers (ViTs)-LLM configurations\cite{icl_vit1, inverge_vit2, mlrg_vit3, ren2gpt_vit4, maira}. Beyond the standard encoder-decoder paradigm, various lines of research have introduced architectural modifications and auxiliary techniques to achieve logical coherence and factual correctness by integrating memory-driven modules\cite{mdt_mem1, csamdt_mem2, mdtm_mem3}, retrieval methods\cite{gu2025radalign} with contrastive learning\cite{dart_ret1, bllm_ret2, CXR-RePaiR}, fine-grained lesion features \cite{ffair_frg1, rgrg}, and medical knowledge graphs \cite{wrrg_gra1, ppked_gra2, kerp_gra3, km_gra4}. More recently, the field is advancing towards native Vision-Language Models (VLMs) \cite{medgem_vlm1, tu2024towards, cclip, medgemini} and Generalist Medical AI (GMAI) \cite{gmai, tu2024towards} architectures. Unlike modular approaches that bridge separate pre-trained vision and language encoders, these foundation models are often trained end-to-end on vast, multimodal datasets. Agent-based frameworks offer a compelling alternative to monolithic LLMs by enabling task decomposition through specialized tools. For instance, a multi-agent architecture was recently proposed to handle disease-specific report generation by assigning specialized tasks to individual agents\cite{mrgagents}. This concept was later extended by incorporating a Concept Bottleneck Model\cite{cbm} to extract critical pathologies; these findings then serve as the foundation for a multi-agent RAG system designed to generate clinically grounded and transparent medical documentation\cite{agentrag}. While Foundational models, LLMs and VLMs excel at producing fluent and descriptive narratives, they fail to generate structured reports grounded in clinical protocols. Though existing agentic frameworks are robust, they often struggle with complex reasoning and factual hallucinations.

\subsection{Explainable AI in Report Generation}

While recent models for automated report generation have achieved strong quantitative performance, their adoption in clinical practice hinges on meeting the critical requirement of explainability. To address this, a prominent line of research has focused on improving visual interpretability\cite{medklip, mgca}. Several studies employed region-guided encoder networks that identify pathological areas and use these localized detections as grounded cues for generation \cite{rgrg, attri_gen, maira2}. Other approaches integrated radiologists' eye-gaze data as a supervisory signal to train more explainable image encoders \cite{gaze_guide}. Further enhancing this paradigm, methods with learnable query tokens optimized visual processing while yielding interpretable spatial attention maps through cross-attention mechanisms \cite{meformer, abnorm_guide}. In parallel, a distinct set of approaches prioritized concept-level explainability\cite{pmrg, gu2025radalign} over spatial heatmaps. These methods output discrete, scalar intermediate representations, to construct more informative prompts for an LLM-based report generator. Extending this idea, other works explicitly extracted key clinical findings from reports using frameworks like RadGraph \cite{radgraph} and leveraged these structured entities to guide the generation process \cite{wang2022medical, dyn_graph}. With the exception of PromptMRG\cite{pmrg}, existing methodologies generally operate under the implicit assumption that pathologies not explicitly documented in the reference reports are absent. Furthermore, all of these approaches rely on general purpose LLMs to extract clinical concepts, which lack verification by radiologists.

\section{Methodology}
\label{section:methodology}

\begin{figure}[!t]
    \centering
    \includegraphics[width=1\textwidth]{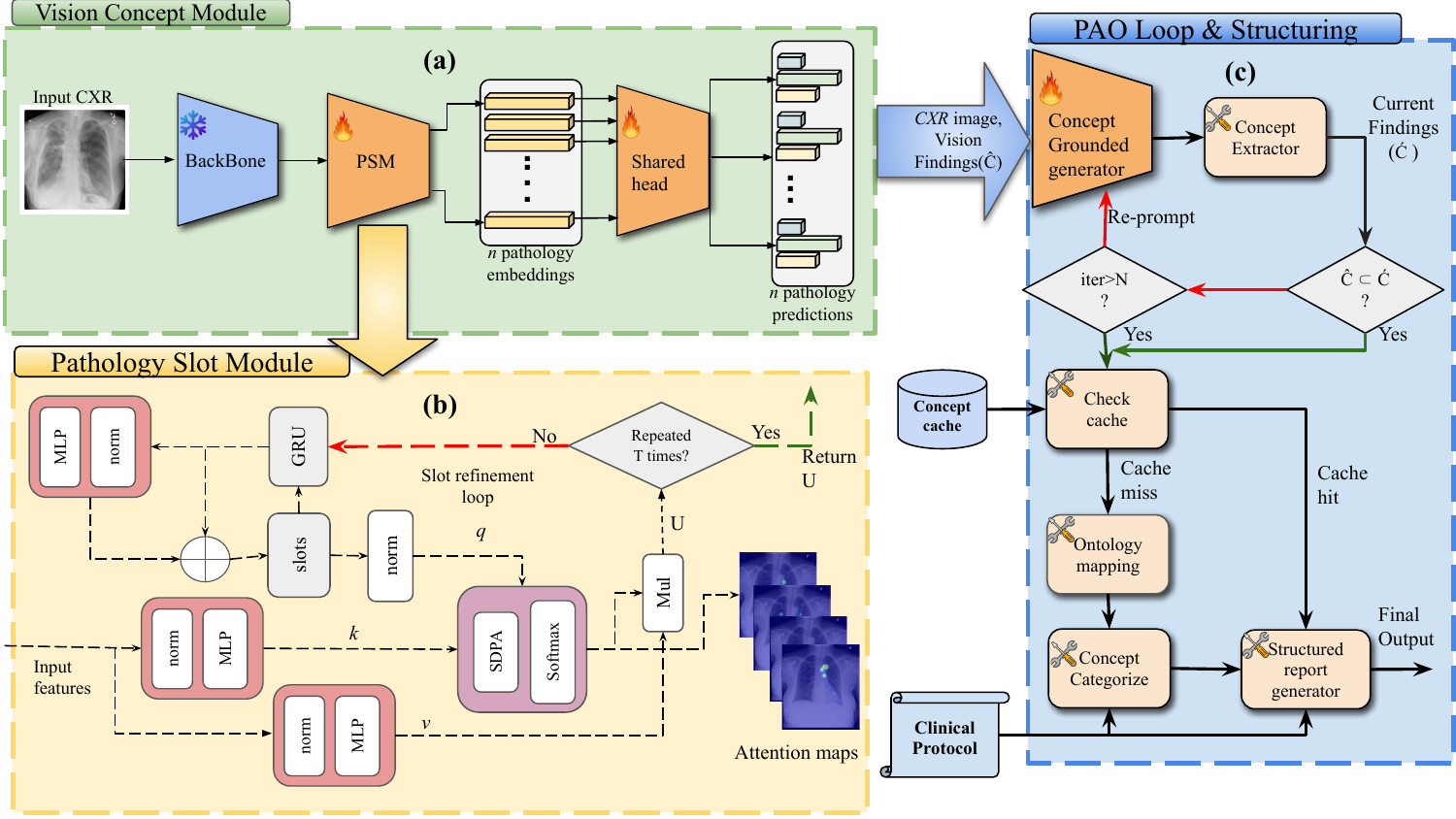}
    \caption{\textbf{Overview of proposed MedPAO-VL architecture.} It mainly comprises of (a)Vision Concept Module for extracting the clinical findings, (b)Pathology slot module for extracting explainable pathology embeddings and (c)An iterative PAO loop to handle report correction and Structuring.}
    \label{fig2}
\end{figure}

We introduce MedPAO-VL, an interpretable, multimodal agent designed to generate protocol-compliant structured radiology reports through an iterative refinement process. Extending the architecture established in MedPAO\cite{medpao}, our system integrates three additional components, the Vision Concept Module, Concept Grounded Report Generator and an Agentic PAO(Plan-Act-Observe) loop. The overall system is as shown in Fig.\ref{fig2}. An overview of tools used in MedPAO architecture can be seen in Supplementary material.

\subsection{Vision Concept Module}
\label{subsec:VisionConceptModule}

\subsubsection{Mitigating Implicit Negative Bias}
Radiology reports frequently omit unresolved or expected findings, reflecting clinical prioritization rather than documented absence~\cite{clinical_uncertainty}. However, prominent AI frameworks~\cite{gloria, mgca} typically enforce a \textit{closed-world assumption} by applying \textit{U-Zeros} strategies~\cite{chexpert} to map unmentioned or \textit{"uncertain"} annotations to definite negatives. This introduces an \textit{implicit negative bias} that corrupts training signals, suppresses uncertainty, and distorts spatial grounding. To eliminate this bias, we formulate pathology prediction as a three-class problem. Given a CXR image $x$, the model predicts a label from $\{\textit{present}, \textit{absent}, \textit{uncertain}\}$ for each of $n$ target pathologies. This tripartite scheme mirrors clinical differential diagnosis by explicitly distinguishing true absence from reporting omission. Fig. \ref{fig2}(a) represents an overview of proposed vision model.


\subsubsection{Explainable Pathology Slot Module}


The Pathology Slot Module (PSM) is motivated by object-centric learning~\cite{slot_attn, ctrlo} and slot attention-based approaches for explainable image classification~\cite{scouter}. Given backbone feature representations, the PSM instantiates $n$ pathology-specific slots, each designed to extract features relevant to its corresponding pathology. For an input feature map $F \in \mathbb{R}^{m \times d}$, where $m$ is the number 
of spatial tokens and $d$ is the feature dimension, the PSM disentangles the 
global representation into $n$ pathology-specific \textit{context vectors} 
$U \in \mathbb{R}^{n \times d}$ via iterative slot refinement over $T$ iterations, 
with slots initialized as $S_0 \sim \mathcal{N}(0, 1) \in \mathbb{R}^{n \times d}$. 
The full forward pass is illustrated in Fig.~\ref{fig2}(b).

\paragraph{\textbf{Key Attention Design.}}
The central contribution of the PSM lies in its attention mechanism. At each 
iteration $t$, pathology-specific slots $S_{t-1}$ attend over the spatial 
feature map to compute attention logits via scaled dot-product attention:
\[
    A_{\text{logits}} = \frac{Q K^\top}{\sqrt{d}} \in \mathbb{R}^{n \times m},
    \qquad
    A_t = \text{softmax}\!\left(\frac{A_{\text{logits}}}{\tau},\ \text{dim}=-1\right),
\]
where $Q$ and $K$ are projections of the current slots and input features 
respectively, and $\tau$ is a temperature hyperparameter. Crucially, unlike the 
original Slot Attention~\cite{slot_attn}, which applies softmax \textit{over the 
slot dimension} to enforce competition between slots per patch, we apply softmax 
\textit{over the patch dimension}. This design choice is motivated by the 
multi-label nature of pathology prediction: a single image patch may carry 
evidence for multiple co-occurring conditions, and forcing slot competition 
would suppress these overlapping signals. Beyond enabling multi-label 
recognition, this formulation yields a direct interpretability benefit - 
each $A_t \in \mathbb{R}^{n \times m}$ constitutes a \textit{per-pathology 
spatial attention map} that is an inherent product of the forward pass, 
providing faithful, pathology-specific visual grounding without any 
post-hoc approximation. The slot state is subsequently updated via a GRU 
cell, integrating the attended context with prior slot state:
\[
    U_t = A_t V \in \mathbb{R}^{n \times d},
    \qquad
    S_t = \text{GRU}(U_t,\ S_{t-1}).
\]
After $T$ iterations, the final context vectors $U_T \in \mathbb{R}^{n \times d}$ 
encode semantically rich, pathology-specific representations for efficient downstream three-class classification task.


\subsubsection{Shared Classification Head.}
A single shared head operates on the final context vectors $U_T$ from the PSM and performs three-class (\textit{present}, \textit{absent}, \textit{uncertain}) prediction for each pathology $i$:
\[
    p^{i} = \mathrm{softmax}\!\left(
        \mathrm{Dropout}\!\left(
            \mathrm{ReLU}(U_T^{i} W_1 + b_1)
        \right) W_2 + b_2
    \right),
\]
yielding $\text{Probability} = \{ p^{1}, \ldots, p^{n} \}$. Rather than $n$ 
independent classifiers, this parameter-efficient design enforces a common 
decision boundary, placing greater representational burden on the PSM while incentivising the slot vectors to be maximally discriminative.


\subsubsection{Interpretability-Aware Training Objective}

The primary training objective is the sum of cross-entropy losses across 
all $n$ pathologies:
\[
    \mathcal{L}_{\text{cls}} = \sum_{i=1}^{n} 
    \text{CrossEntropy}\!\left(p^{i},\, gt^{i}\right),
\]
where $p^i$ and $gt^i$ denote the predicted probabilities and ground-truth 
label for pathology $i$.

\paragraph{\textbf{Similarity Loss as an Interpretability Regularizer.}}
Optimizing $\mathcal{L}_{\text{cls}}$ alone is insufficient for producing 
meaningful explanations. Since the PSM applies softmax over the patch 
dimension, the slot attention maps are susceptible to \textit{representational 
collapse}, converging toward near-identical distributions across all $n$ 
slots. To explicitly regularize against this failure mode and enforce 
\textit{per-pathology spatial diversity}, we introduce $\mathcal{L}_{\text{sim}}$, 
a redundancy penalty over the attention maps. Following~\cite{LabelFreeCB}, each normalized attention map $A_k$ is $\ell_2$-normalized along the patch dimension, and a pairwise cos-cubed similarity matrix $M \in \mathbb{R}^{n \times n}$ is 
computed by element-wise cubing the normalized maps and taking their inner product:
\[
    \hat{A}_{k} = \frac{A_{k}}{\lVert A_{k} \rVert_2} \in \mathbb{R}^{n \times m},
    \qquad
    M = \hat{A}^{\,3}\!\left(\hat{A}^{\,3}\right)^{\!\top},
\]
where $\hat{A}^3$ denotes element-wise cubing, and each entry $M_{i,j}$ 
measures the cos-cubed similarity between the attention maps of pathology 
$i$ and pathology $j$. The cubic transformation amplifies high-similarity 
signals while suppressing noise, making the penalty sensitive to genuine 
redundancy rather than incidental overlap. To penalize only redundancy 
while permitting benign anti-correlation between anatomically distinct 
pathologies, ReLU is applied exclusively to off-diagonal entries:
\[
    \mathcal{L}_{\text{sim}} 
    = \frac{1}{n(n-1)}
    \sum_{i=1}^{n}\sum_{\substack{j=1\\j \neq i}}^{n}
    \mathrm{ReLU}(M_{i,j}).
\]

The complete training loss is a linear combination of both terms:
\[
    \mathcal{L} = \mathcal{L}_{\text{cls}} + \lambda_{\text{sim}}\,
    \mathcal{L}_{\text{sim}},
\]
where $\lambda_{\text{sim}}$ balances classification performance against 
interpretability quality. The result is a training objective 
that jointly optimizes for diagnostic accuracy and spatial interpretability, 
ensuring that the PSM's explanations are both clinically faithful and 
visually diverse across pathologies. For further analysis of ${L}_{\text{sim}}$, refer to Supplementary material.

\subsection{Concept Grounded Report Generator}
\label{subsec:GroundedReportGenerator}
The Concept Grounded Report Generator is a multimodal model tasked at synthesizing a fluent, unstructured report. Unlike unconstrained generation, this module adopts a concept-driven approach: it receives both the CXR image and semantic constraints, a list of present and absent clinical concepts from the Vision Concept Module and renders these findings into medically coherent report. This dual-input design serves two complementary purposes. First, by grounding the generation on a curated concept vocabulary rather than raw image features alone, the outputs become traceable to discrete, verifiable clinical findings. This traceability is a critical prerequisite for clinical trustworthiness, allowing practitioners to audit the specific concepts driving any narrative statement. Second, utilizing the image as an active visual reference provides the necessary multimodal context to produce naturally expressive descriptions, avoiding the disjointedness of mere concept concatenation.

\subsection{The Agentic PAO Loop}
\label{subsec:AgenticPAOLoop}

The PAO loop shown in Fig.\ref{fig2}(c) is an iterative loop that adapts to the complexity of the clinical findings, by offering the following capabilities:

\textbf{Self-Correction Loop and Report Structuring.}
After extracting the visual findings, the agent begins by initiating the \textit{Concept Grounded Report Generator} to produce an initial findings narrative. To ensure clinical fidelity, the Observe phase employs Semantic Overlap Checker. This module extracts clinical entities from the generated narrative and computes the semantic similarity against the concept list $\hat{C}$ from the Vision Concept Module. If the checker identifies a discrepancy, the agent triggers a Re-Plan and Act cycle. In this phase, the generator is re-prompted to refine the specific section containing the discrepancy. This loop continues for up to N iterations or until the vision concept list $\hat{C} \subseteq \text{CurrentFindings}$. After loop termination, the agent uses various tools from MedPAO like concept extraction, ontology mapping, concept categorizer and structured report generator to structure the free-style report according to a given clinical protocol. Refer to Supplementary for an end to end example of MedPAO-VL.

\textbf{Amortized Efficiency via Concept Caching.} To mitigate the computational overhead of iterative agentic loops, the PAO framework employs a Dynamic Concept Cache for ABCDEF protocol categorization. By storing and retrieving mappings for frequently recurring CXR findings, the agent bypasses high-latency \texttt{Ontology Mapping}, and \texttt{Categorizer} tool invocations on cache hits. New mappings are dynamically cached, allowing the system to reuse prior knowledge and optimize inference (see Supplementary for further details on PAO implementation).

\section{Experiments}
\label{sec:experiment}

\subsection{Dataset \& Pre-processing} 
\label{subsec:data_curation}

We utilize a subset of MIMIC-CXR\cite{mimic_cxr} to train both the Vision Concept Module and the Concept Grounded Report Generator. We selected studies where the reports explicitly contain \texttt{HISTORY}, \texttt{FINDINGS}, and \texttt{IMPRESSION} sections, yielding approximately 53,000 CXR–report pairs from the train set. The \texttt{FINDINGS} section serves as the primary input to our concept extraction pipeline, which identifies salient clinical findings from free-form text. We leveraged the same concept extraction pipeline from ~\cite{medpao} which was validated on a gold standard dataset annotated by radiologists. Since a same observation may be expressed in many linguistic forms, the extracted concepts are normalized to canonical SNOMED CT ontology using the medical entity normalization model~\cite{sapbert}, enabling uniform grouping across heterogeneous representations of similar findings. An LLM (DeepSeek-R1-Distill-Llama-70B~\cite{DeepSeekR1}) is used to separate the present and absent concepts within each group. After radiologists validation, low-frequency or incoherent groups are filtered out, producing a final vocabulary of 236 unique, standardized concept groups. Additional details on radiologist validation is in Supplementary material. At the end of pre-processing steps, the curated dataset comprises approximately 50,000 (CXR, report, concept) triplets that form the basis for all models presented in this work. We utilize MIMIC-CXR\cite{mimic_cxr} test set and IU-XRAY\cite{iu-xray} test set for evaluation and the same pre-processing pipeline is applied on the test set to obtain ground-truth concepts from radiology reports, against which all models are validated. Additional details on data curation are provided in Supplementary material.

\subsection{Implementation Details}

For training the \textbf{Vision Concept Module}, the dataset obtained in sub-section \ref{subsec:data_curation} is partitioned into a 90:10 train–validation split using stratified sampling on the \textit{clear lungs} pathology. We use the MedSigLIP vision encoder~\cite{medgem_vlm1} as a fixed backbone, operating on an image resolution of 448x448. The PSM block produces 236 pathology-specific slots corresponding to the 236 target concepts. The similarity loss weight is set to $\lambda_{\text{sim}} = 3$. We train for 50 epochs using AdamW with a cosine learning rate schedule, an initial learning rate of $5\times10^{-5}$, and a batch size of 32. For the \textbf{Concept Grounded Report Generator}, we finetune MedGemma 4B~\cite{medgem_vlm1} on 10,000 samples for ablation studies and utilize the best hyperparameters to scale the training over 50,000 samples. All the experiments were conducted on a single NVIDIA A6000 (48 GB) GPU. Further implementation details are provided in Supplementary material.


\subsection{Report Generation Evaluation}

\subsubsection{Unstructured Report Evaluation}

As shown in Table~\ref{tab:results_comparison} and Table~\ref{tab:model_detailed_metrics}, the proposed Concept Grounded Report Generator sets a new benchmark for clinical accuracy in chest X-ray report generation. Our 10K and 50K variants, which represent the report generator fine-tuned on 10,000 and 50,000 samples respectively, demonstrate the scalability of our approach. Specifically, the 50K variant achieves a state-of-the-art CheXbert-F1\cite{chexbert} of 0.617 and a RadGraph-F1\cite{radgraph} of 0.427 on the MIMIC-CXR dataset, substantially outperforming significantly larger models such as MAIRA-2\cite{maira2} and Flamingo-CXR\cite{flamingocxr}. This superior performance in clinical metrics, paired with a high CheXbert Micro-F1 of 0.706, underscores the model’s ability to correctly identify pathologies and map complex anatomical-clinical relationships. Furthermore, even the 10K variant remains competitive against SOTA baselines, validating the effectiveness of our radiologist-validated concept standardization even when fine-tuned on a smaller subset of data (compared to 1M samples in MAIRA-2 and 500K in Flamingo-CXR). While our model remains competitive in NLG metrics, the high GREEN\cite{green_metric} mean scores (0.733 for IU X-RAY) highlight that our generated reports maintain high clinical utility and diagnostic integrity, effectively minimizing critical omissions compared to existing generative approaches.

\begin{table}
\centering
\caption{\textbf{Performance Comparison on MIMIC-CXR and IU X-RAY Datasets.} This table presents the performance of various models against our Concept Grounded report Generator on generating unstructured radiology reports(without PAO loop) for the MIMIC-CXR and IU X-RAY test sets. The Sections column indicates whether the model generates the \texttt{FINDINGS}(F) or \texttt{IMPRESSION}(I) section of the report, with metrics sourced from published research.}
\label{tab:results_comparison}
\resizebox{\textwidth}{!}{
\begin{tabular}{@{}llcccccc@{}}
\toprule
\multirow{2}{*}{\textbf{Dataset}} & \multirow{2}{*}{\textbf{Methods}} & \multirow{2}{*}{\textbf{Year}} & \multirow{2}{*}{\textbf{Section}} & \multicolumn{2}{c}{\textbf{NLG Metrics}} & \multicolumn{2}{c}{\textbf{Clinical Metrics}} \\ \cmidrule(lr){5-6} \cmidrule(lr){7-8}
 &  & & & BLEU-4 & ROUGE-L & CheXbert-F1 & RadGraph-F1 \\ \midrule
\multirow{16}{*}{MIMIC-CXR} 
 & R2Gen\cite{mdt_mem1} & 2020 & F+I & 0.103 & 0.277 & 0.276 & 0.134 \\
 & CXR-RePaiR\cite{CXR-RePaiR} & 2021 & F & 0.021 & 0.143 & 0.274 & 0.091 \\
 & $\mathcal{M}^2$ Transformer\cite{m2transformer} & 2021 & F & 0.114 & - & - & 0.220 \\
 & RGRG\cite{rgrg} & 2023 & F & 0.126 & 0.264 & 0.447 & - \\
 & METransformer\cite{meformer} & 2023 & F & 0.124 & 0.291 & 0.311 & - \\
 & R2GenGPT\cite{ren2gpt_vit4} & 2023 & F+I & 0.134 & 0.297 & 0.389 & - \\
 & Flamingo-CXR\cite{flamingocxr} & 2023 & F+I & 0.101 & 0.297 & 0.519 & 0.205 \\
 & Med-PaLM M, 84B\cite{tu2024towards} & 2024 & F & 0.113 & 0.273 & - & 0.267 \\
 & Med-Gemini-2D\cite{medgemini} & 2024 & F+I & 0.205 & 0.283 & - & 0.244 \\
 & PromptMRG\cite{pmrg} & 2024 & F & 0.112 & 0.268 & 0.476 & - \\
 & InVERGe\cite{inverge_vit2} & 2024 & F & 0.100 & 0.309 & - & - \\
 & LLaVA-Rad\cite{llavarad} & 2024 & F+I & 0.154 & 0.306 & 0.395 & 0.294 \\
 & Chexagent\cite{chexagent} & 2024 & F+I & - & \textbf{0.450} & 0.444 & 0.266 \\
 & MAIRA-2\cite{maira2} & 2024 & F & \textbf{0.231} & 0.384 & 0.416 & 0.346 \\
 & MedGemma, 4B\cite{medgem_vlm1} & 2025 & F+I & 0.019 & 0.131 & - & 0.295 \\
\rowcolor{blue!10} \cellcolor{white} & Ours (10K) &  & F & 0.102 & 0.298 & 0.563 & 0.376 \\
\rowcolor{blue!10} \cellcolor{white} & Ours (50K) &  & F & 0.136 & 0.328 & \textbf{0.617} & \textbf{0.427} \\ \midrule

\multirow{10}{*}{IU X-RAY} 
 & R2Gen\cite{mdt_mem1} & 2020 & F+I & 0.165 & 0.371 & 0.136 & - \\
 & $\mathcal{M}^2$ Transformer\cite{m2transformer} & 2021 & F & 0.131 & - & - & - \\
 & RGRG\cite{rgrg} & 2023 & F & 0.063 & 0.180 & 0.180 & - \\
 & R2GenGPT\cite{ren2gpt_vit4} & 2023 & F+I & \textbf{0.173} & \textbf{0.377} & - & - \\
 & PromptMRG\cite{pmrg} & 2024 & F & 0.112 & 0.268 & \textbf{0.476} & - \\
 & InVERGe\cite{inverge_vit2} & 2024 & F & 0.100 & 0.309 & - & - \\
 & MAIRA-2\cite{maira2} & 2024 & F & 0.117 & 0.274 & 0.319 & 0.271 \\
\rowcolor{blue!10} \cellcolor{white} & Ours (10K) &  & F & 0.083 & 0.327 & 0.203 & 0.342 \\
\rowcolor{blue!10} \cellcolor{white} & Ours (50K) &  & F & 0.097 & 0.341 & 0.424 & \textbf{0.401} \\ \bottomrule
\end{tabular}
}
\end{table}

\begin{table}
\centering
\caption{\textbf{Comprehensive Evaluation Metrics of Concept Grounded report Generator (50K variant, without PAO loop) on MIMIC-CXR and IU X-RAY test-sets}}
\label{tab:model_detailed_metrics}
\resizebox{\textwidth}{!}{
\begin{tabular}{@{}lcccccccccccccc@{}}
\toprule
\multirow{2}{*}{\textbf{Dataset}} & \multicolumn{6}{c}{\textbf{NLG}} & \multicolumn{2}{c}{\textbf{CheXbert}} & \multicolumn{2}{c}{\textbf{GREEN}} & \multicolumn{3}{c}{\textbf{RadGraph}} \\ \cmidrule(lr){2-7} \cmidrule(lr){8-9} \cmidrule(lr){10-11} \cmidrule(lr){12-14} 
 & BLEU-1 & BLEU-2 & BLEU-3 & BLEU-4 & METEOR & ROUGE-L & Micro-F1 & Macro-F1 & Mean & STD & $\text{RG}_\text{E}$ & $\text{RG}_\text{ER}$ & $\overline{\text{RG}}_\text{ER}$ \\ \midrule
MIMIC-CXR & 0.431 & 0.272 & 0.188 & 0.136 & 0.407 & 0.328 & 0.706 & 0.617 & 0.472 & 0.214 & 0.403 & 0.374 & 0.303 \\
IU X-RAY  & 0.416 & 0.239 & 0.150 & 0.097 & 0.469 & 0.341 & 0.619 & 0.424 & 0.733 & 0.225 & 0.495 & 0.458 & 0.375 \\ \bottomrule
\end{tabular}
}
\end{table}

\begin{table}
\centering
\caption{\textbf{Radiologist evaluation of structured reports with $\pm 95\%$ CI}}
\label{tab:radiologist_evals}
\begin{tabular}{lcccc}
\hline
 & \multicolumn{2}{c}{\textbf{MedGemma 27B}} & \multicolumn{2}{c}{\textbf{MedPAO-VL}} \\
\cmidrule(lr){2-3} \cmidrule(lr){4-5}
 & Accuracy score & Structure score & Accuracy score & Structure score \\
\hline
Radiologist1 & $2.88 \pm 0.44$  & $3.84 \pm 0.36$  & $3.40 \pm 0.32$ & $4.44 \pm 0.24$ \\
Radiologist2 & $3.48 \pm 0.56$ & $4.12 \pm 0.28$ & $3.60 \pm 0.40$ & $4.22 \pm 0.32$ \\
\midrule
\textbf{Combined Mean} & $3.18 \pm 0.36$ & $3.98 \pm 0.24$ 
                       & $\mathbf{3.50 \pm 0.28}$ 
                       & $\mathbf{4.28 \pm 0.20}$ \\
\hline
\end{tabular}
\end{table}

\subsubsection{Structured Report Evaluation}

To evaluate the quality of the generated structured radiology reports, we defined two primary scoring metrics: (1) an \textit{Accuracy Score}, which measures the clinical correctness of the generated findings, and (2) a \textit{Structural Score}, which assesses the extent to which the generated reports conform to the predefined reporting protocol. The evaluation was conducted by two senior radiologists on a balanced subset of 50 cases drawn from the MIMIC-CXR test set. To ensure a representative evaluation, these cases were selected via stratified random sampling, which preserved the underlying distribution of both CheXpert labels\cite{chexpert} and report complexity (utilizing an \textit{is complex} flag for reports containing $\ge 9$ concepts; refer to Supplementary material). For each case, the radiologists were presented with samples consisting of reports generated by our agent and reports produced by the MedGemma-27B multimodal model, along with the corresponding CXR image and the ground-truth report. To ensure an unbiased assessment, the radiologists underwent a blinded evaluation, without knowledge of the method responsible for generating the report. The mean evaluation results across all assessed cases are reported in Table~\ref{tab:radiologist_evals}. Full inter-rater statistics, score distributions and the scoring criteria are described in Supplementary material. 


\subsection{Interpretability and Representational Analysis}

\subsubsection{Latent Orthogonality}
\begin{figure*}[!t]
    \begin{subfigure}[b]{0.5\textwidth}
        \centering
        \includegraphics[trim=3cm 0cm 3cm 0cm, clip,width=1\linewidth]{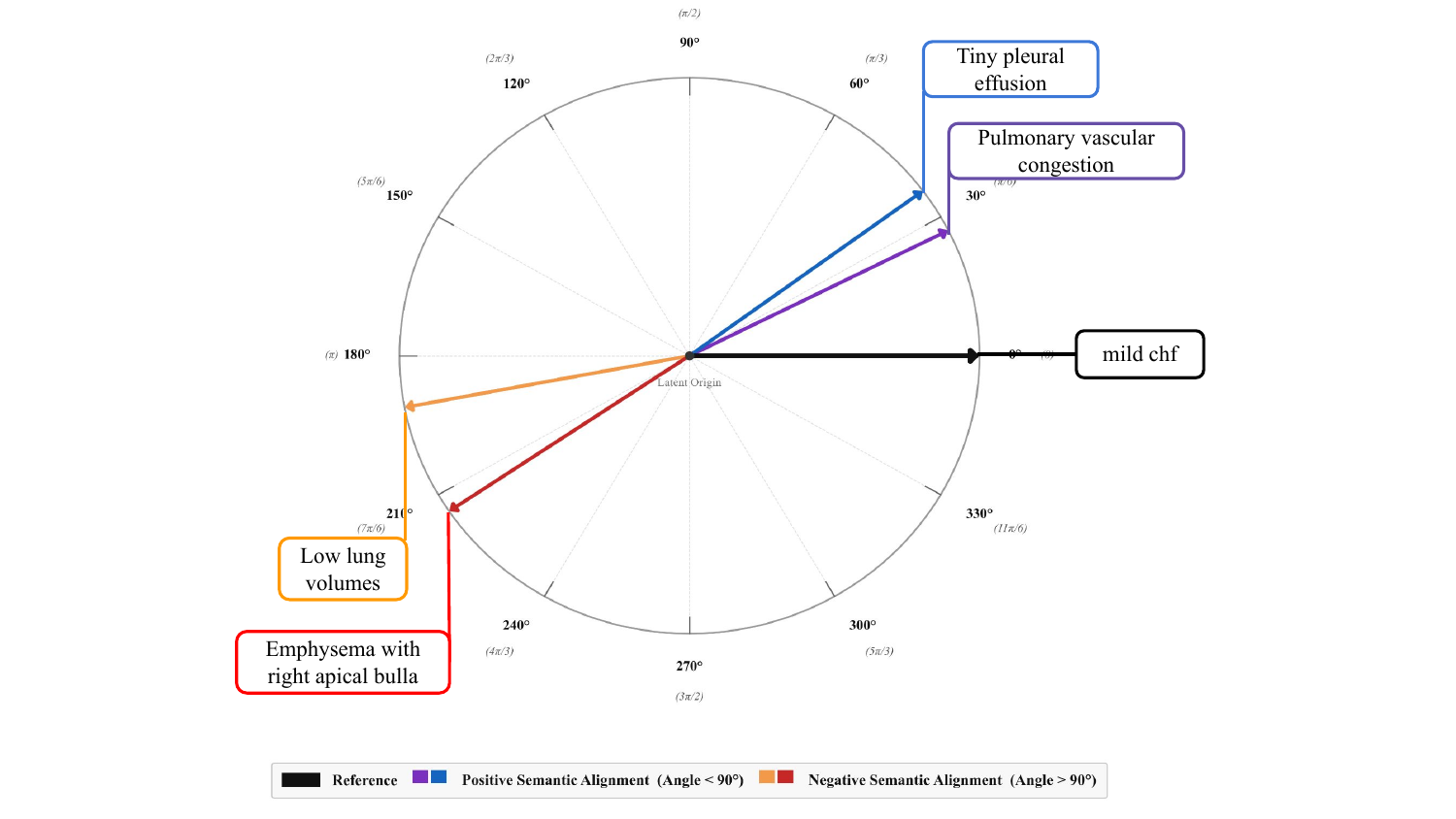}
        \caption{}
        \label{fig:chf}
    \end{subfigure}
    \hfill
    \begin{subfigure}[b]{0.5\textwidth}
        \centering
        \includegraphics[trim=3cm 0cm 3cm 0cm, clip,width=1\linewidth]{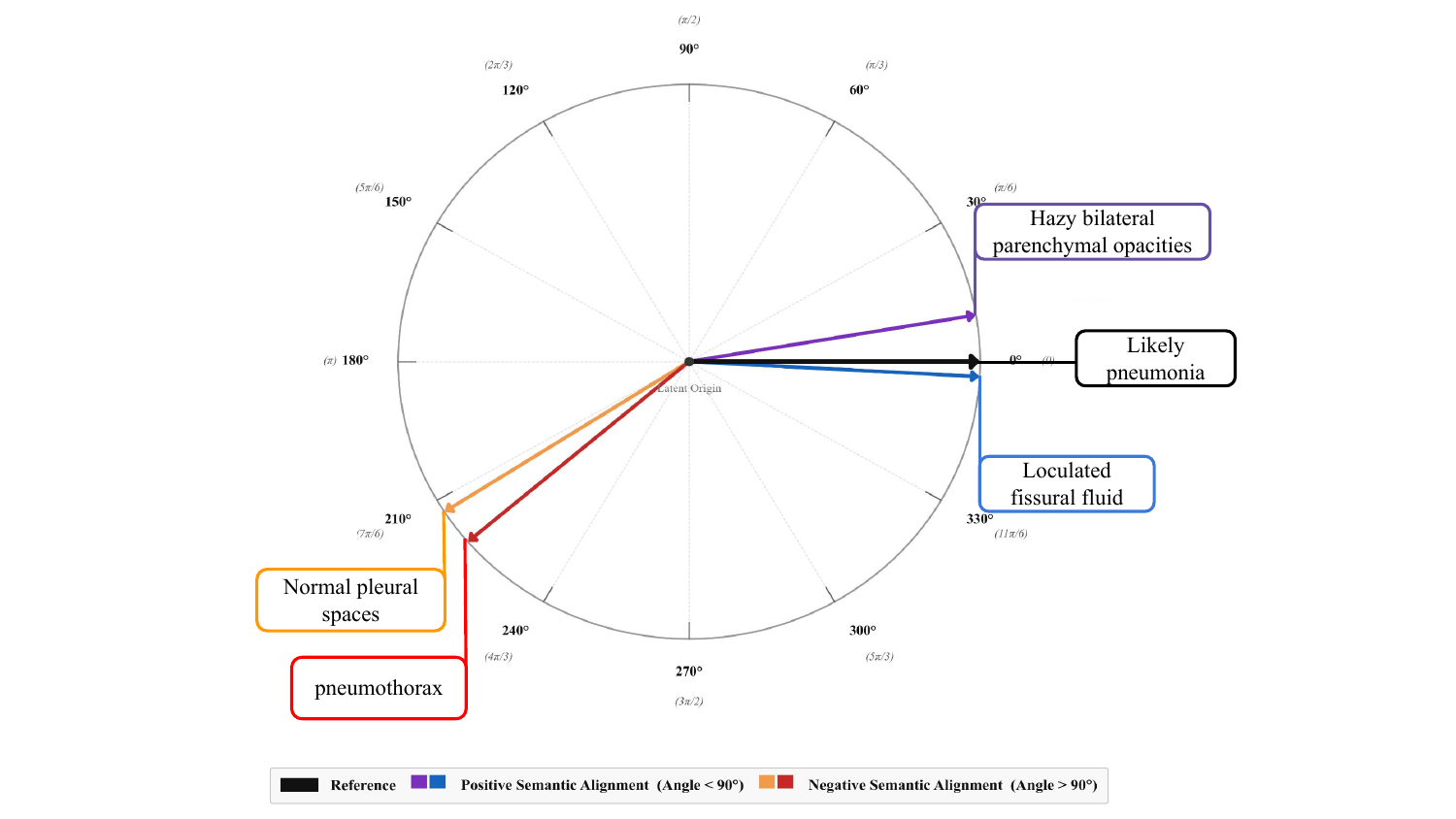}
        \caption{}
        \label{fig:pneumonia}
    \end{subfigure}
    \hfill
    
    \caption{\textbf{Latent pathology slot vector similarity.} 2D cosine similarity projections for (a) ``mild CHF'' and (b) ``likely pneumonia'' and related pathologies.}
    \label{fig:macro_micro_sim}
\end{figure*}

\begin{figure}[!t]
    \centering
    \includegraphics[width=0.9\textwidth]{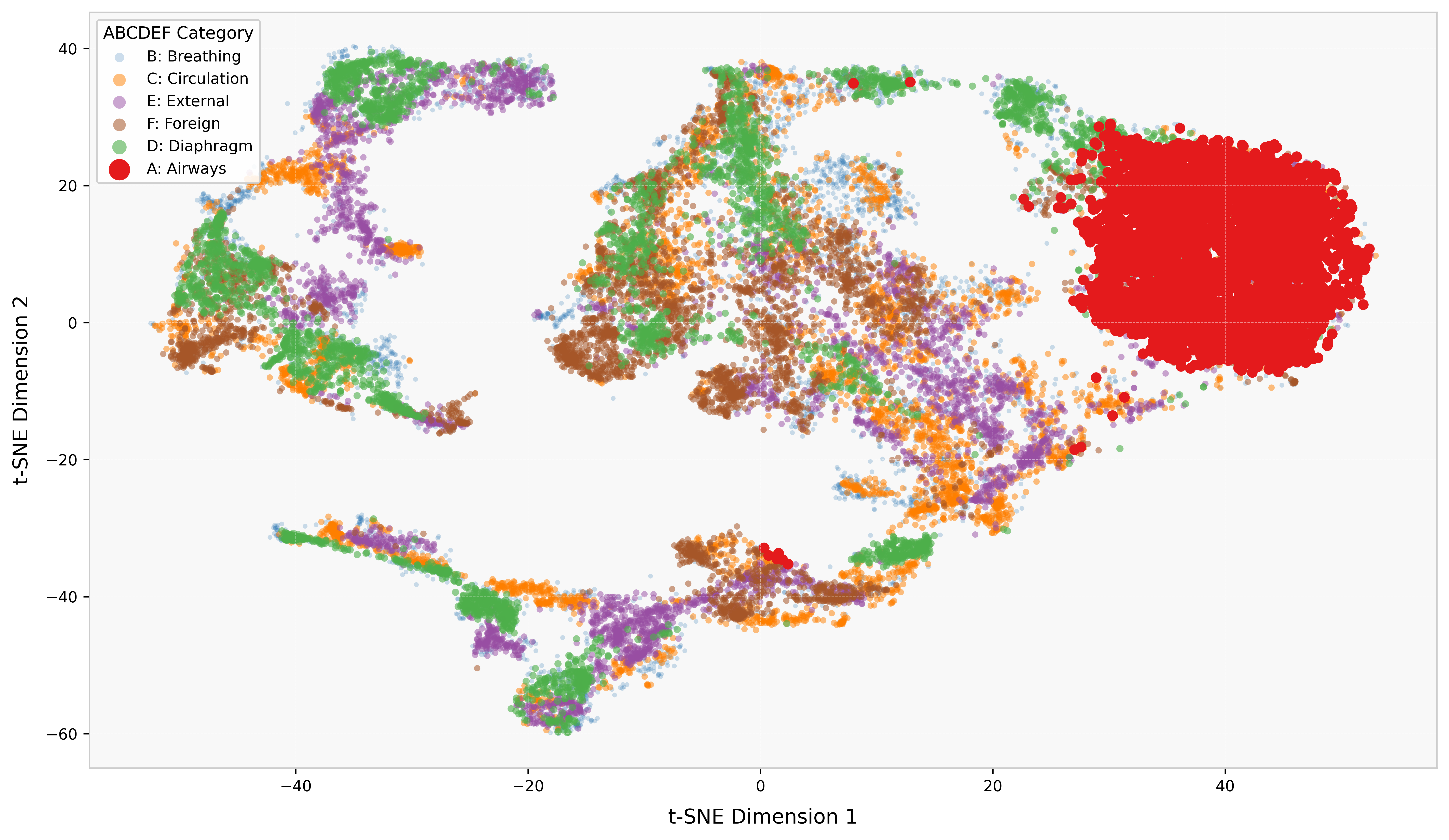}
    \caption{t-SNE visualization of the slot activation space against ABCDEF clinical protocol.}
    \label{fig:tsne_abcd}
\end{figure}

To evaluate the representational fidelity of the PSM, we analyze inter-slot cosine similarity of the final context vectors $U_T$. Context vectors are unit-normalized and projected onto a 2D plane anchored to a reference pathology, revealing clinically coherent directional structure: related findings cluster near the reference axis while opposing pathologies are displaced into the negative half-space. For \textit{likely pneumonia} (Fig.\ref{fig:pneumonia}), \textit{hazy bilateral parenchymal opacities} ($\cos{=}{+}0.985$, $9.7^\circ$) and \textit{likely loculated fissural fluid} ($\cos{=}{+}0.985$, $3.1^\circ$) align tightly with the reference, consistent with classical radiographic presentation, whereas \textit{pneumothorax} ($\cos{=}{-}0.281$) and \textit{normal pleural surfaces} ($\cos{=}{-}0.348$) are geometrically opposed. For \textit{mild CHF} (Fig.\ref{fig:chf}), \textit{mild pulmonary vascular congestion} ($\cos{=}{+}0.893$) and \textit{tiny pleural effusion} ($\cos{=}{+}0.696$) project positively as expected sequelae of elevated hydrostatic pressure, while \textit{emphysema with right apical bulla} ($\cos{=}{-}0.360$) and \textit{low lung volumes} ($\cos{=}{-}0.377$) are displaced into the opposing quadrant, reflecting their association with obstructive rather than congestive physiology. These geometric relationships demonstrate that the PSM encodes clinically meaningful differentials as directional structure in embedding space. For further analysis on clinical co-occurrence of learnt pathology embeddings refer to Supplementary material.

\begin{figure}[!t]
    \centering
    \includegraphics[trim=0cm 0cm 0cm 8.1cm, clip,width=1\linewidth]{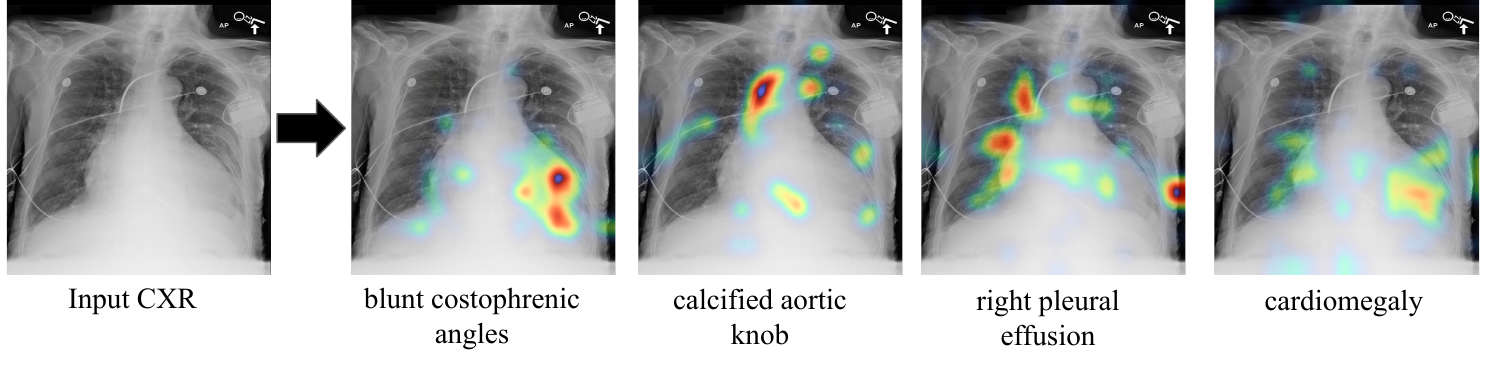}
    \caption{Input CXR and Explainable Attention maps from PSM block}
    \label{fig:attn_maps}
\end{figure}

\subsubsection{Emergent Anatomical Organization in Latent Manifold}
We evaluated the representational fidelity of the PSM using t-SNE over the 236-dimensional slot activation space. A significant emergent property is observed: while the model was never trained to categorize findings into the ABCDEF protocol, the latent vectors naturally organize into clusters that coincide with these anatomical categories. To ensure a balanced representation, we utilized a stratified subset of 4,000 slot instances per category from the MIMIC-CXR test set. We employed a cosine distance metric for the projection to prioritize vector similarity over absolute magnitude, effectively capturing the diagnostic relationships within the high-dimensional manifold. As shown in Fig.\ref{fig:tsne_abcd}, the distinct spatial separation between Airways (A), Breathing (B), and Circulation (C) clusters validates that our concept-driven architecture disentangles co-occurring pathologies based on their anatomical origin. This alignment provides a verifiable diagnostic manifold that mirrors radiological reasoning, bridging the gap between deep vision features and clinical interpretability. Extended qualitative analysis of per-slot tsne plot can be found in Supplementary material.

\subsubsection{Spatial Explainability}
As discussed in Section~\ref{subsec:VisionConceptModule}, the PSM block is 
designed not only to capture pathology-specific representations but also to 
generate interpretable attention maps as an inherent product of its forward 
pass, requiring no post-hoc approximation. Fig.~\ref{fig:attn_maps} 
illustrates this capability on a representative complex CXR containing four 
co-occurring pathologies. Crucially, each slot attends to a distinct and 
anatomically faithful region: the \textit{blunt costophrenic angles} slot 
correctly localizes attention to the lower lung bases where pleural 
blunting is evident; the \textit{calcified aortic knob} slot concentrates 
activation at the upper mediastinum near the aortic arch; the \textit{right 
pleural effusion} slot exhibits a lateralized response confined to the right 
lower hemithorax; and the \textit{cardiomegaly} slot distributes attention 
broadly across the cardiac silhouette, consistent with diffuse enlargement.

\section{Ablations}
\label{sec:ablations}

\begin{table}[ht]
\centering
\caption{Evaluation results of Vision Concept module from its baseline model to the proposed architecture}
\label{tab:ablation_vision}
\resizebox{\linewidth}{!}{
\begin{tabular}{llcccccc}
\toprule
\textbf{Dataset} & \textbf{Methods} & \textbf{Trainable Params.} & \textbf{AUC} & \textbf{F1-score} & \textbf{Precision} & \textbf{Recall} \\
\midrule
\multirow{4}{*}{MIMIC-CXR} 
& Baseline model & 2.7M & 0.64 & 0.485 & 0.42 & 0.574  \\
& Baseline + PSM &  47.2M & 0.679 & 0.484 & 0.41 & 0.59  \\
& Baseline + PSM + Shared Classifier Head & 6.1M & 0.77 & 0.467 & 0.44 & 0.497  \\
\rowcolor{blue!10} \cellcolor{white} & Proposed(Baseline + PSM + Shared Classifier + Similarity loss) & 6.1M & 0.784 & 0.48 & 0.478 & 0.567  \\
\bottomrule
\end{tabular}
}
\end{table}

\begin{table}[ht]
\centering
\caption{Evaluation results on the impact of model scale, fine-tuning, and the proposed clinical concept integration in Concept Grounded Report Generator}
\label{tab:ablation_llm}
\resizebox{\linewidth}{!}{
\begin{tabular}{llcccccc}
\toprule
\textbf{Dataset }& \textbf{Methods} & \textbf{BLEU-1 }& \textbf{BLEU-2} & \textbf{BLEU-3} & \textbf{BLEU-4} & \textbf{ROUGE-L }& \textbf{METEOR} \\
\midrule
\multirow{4}{*}{MIMIC-CXR} 
& MedGemma-4B MM zero shot & 0.139 & 0.068 & 0.035 & 0.019 & 0.131 & 0.253 \\
& MedGemma-27B MM zero shot & 0.235 & 0.109 & 0.051 & 0.026 & 0.165 & 0.234 \\
& MedGemma-4B MM Finetuned w/o Concepts (10K) & 0.275 & 0.163 & 0.104 & 0.071 & 0.227 & 0.265 \\
\rowcolor{blue!10} \cellcolor{white} & Proposed (MedGemma-4B Finetuned with Concepts, 10K) & 0.377 & 0.224 & 0.147 & 0.102 & 0.298 & 0.358 \\
\bottomrule
\end{tabular}
}
\end{table}


\begin{table}[ht]
\centering
\small
\caption{Evaluation of the Plan-Act-Observe (PAO) loop across various configurations.}
\label{tab:pao_ablation}
\resizebox{\textwidth}{!}{
\begin{tabular}{lccccccc}
\toprule
\textbf{System Configuration}
    & \textbf{CheXbert-F1} $\uparrow$
    & \textbf{RadGraph-F1} $\uparrow$
    & \textbf{GREEN} $\uparrow$
    & \textbf{$\phi^{+}$ Ratio} $\uparrow$
    & \textbf{Mean Iter.}
    & \textbf{LLM Calls} $\downarrow$
    & \textbf{Time (s)} $\downarrow$ \\
\midrule
Deterministic Baseline
    & 0.617
    & 0.427
    & $0.47 \pm 0.21$
    & 0.602
    & 1.0
    & 6.0
    & 218 \\

PAO Full (Cold Cache)
    & 0.634
    & 0.448
    & $0.61 \pm 0.20$
    & 0.803
    & 2.8
    & 11.2
    & 295 \\

\rowcolor{blue!10}
PAO Full (Warm Cache)
    & 0.634
    & 0.448
    & $0.61 \pm 0.20$
    & 0.803
    & 2.8
    & 9.8
    & 163 \\
\bottomrule
\end{tabular}
}
\end{table}


\subsection{Vision model ablations}
As presented in Table \ref{tab:ablation_vision}, we compare successive stages of the vision model design using standard classification metrics. The \emph{Baseline} configuration corresponds to directly using the vision backbone features followed by 236 independent classification heads for multi-label prediction. Incorporating the PSM module yields a modest improvement over the Baseline while additionally enabling the generation of interpretable attention maps. The PSM variant with a shared classifier further improves performance while reducing the overall parameter count. Finally, introducing the similarity loss has minimal impact on the quantitative metrics but noticeably enhances the quality of the attention maps by encouraging more diverse, pathology-specific focus patterns. 


\subsection{Impact of Concept Grounding}
As shown in Table \ref{tab:ablation_llm}, the finetuned MedGemma-4B model significantly outperforms both the 4B and 27B zero-shot baselines. This result strongly suggests that task-specific adaptation is far more critical for generating high-quality reports than raw model scale alone, making finetuning an essential step. However, the most critical comparison is between the finetuned model without concepts and our complete proposed model. As the results demonstrate, incorporating medical concepts yields performance boost, positioning our proposed model clearly ahead of all other variants. This jump highlights that the concept integration provides the crucial semantic grounding to achieve the highest level of clinical accuracy and fluency.


\subsection{PAO Loop Ablations}
\label{subsec:pao_ablation}

The PAO ablations shown in Table~\ref{tab:pao_ablation} comprises of three different combinations tested on MIMIC-CXR test. For all the experiments the maximum number of refinement steps is set to 3. The deterministic baseline experiment runs all the tools sequentially and achieves a concept coverage ratio($\phi^{+}$) of $0.602$, indicating that approximately $40\%$ of vision module predicted findings are silently omitted from the final freestyle report. Introducing the Self Correction loop(PAO Full, Cold Cache) raises $\phi^{+}$ to $0.803$ and improves ChexBERT-F1\cite{chexbert}, Radgraph-F1\cite{radgraph} and GREEN\cite{green_metric} scores, with a mean iteration count of $2.8$ confirming that the loop triggers genuine re-planning in the large majority of reports rather than terminating in a single pass. While iterative repair increases mean LLM calls from $6.0$ to $11.2$, the warm-cache condition recovers this overhead substantially, by retrieving previously computed concept-to-ABCDEF mappings from the cache. Through caching the frequent findings, inference time is reduced from $295$s to $163$s, a $45\%$ reduction while quality metrics remain unchanged.



\section{Conclusion}
\label{sec:conclusion}

We introduced MedPAO-VL, a transparent, multimodal agent designed to generate protocol-compliant, structured radiology reports. Unlike opaque end-to-end models, MedPAO-VL integrates generative capabilities with verifiable agentic reasoning to yield explicit explainability maps and auditable intermediate reasoning traces. The architecture's three core components - the Vision Concept Module, the Concept Grounded Report Generator, and the iterative PAO loop - work in concert to preserve the inherent uncertainty of clinical observations, provide faithful visual localization, and enforce strict adherence to medical ontologies. Extensive evaluations confirm that our approach produces reliable, factually consistent, and protocol-compliant structured radiology reports. Despite these strengths, the system's current diagnostic reach is constrained by a predefined vocabulary of 236 concept groups, limiting the documentation of ultra-rare anomalies. Future work will address this limitation by dynamically expanding the clinical vocabulary and extending the agentic framework to accommodate multi-modal imaging and broader diagnostic workflows.

\bibliographystyle{splncs04}
\bibliography{main}
\newpage
\appendix
\setcounter{page}{1}
\setcounter{section}{0}
\begin{center}
    \textbf{Supplementary Material}
\end{center}

\section{The ABCDEF Protocol for Systematic CXR Interpretation}
\label{app:abcdef}
Chest radiograph interpretation is a foundational clinical skill that is prone to errors of omission when approached unsystematically. To mitigate this risk, the ABCDEF mnemonic has been widely adopted as a structured, sequential framework for reviewing chest X-rays (CXRs), ensuring that all anatomical regions and radiological domains are evaluated before a diagnostic impression is formed~\cite{abcdef}. Each letter of the mnemonic corresponds to a distinct category of findings, as summarised in Table~\ref{tab:abcdef}.

\paragraph{A~---~Airways.}
Evaluation begins at the midline with the trachea, which should be central at the sternal notch, with a permissible slight rightward deviation at the aortic arch. The trachea is traced inferiorly to the carina, where it bifurcates into the right and left main bronchi; the normal carinal angle is less than $105^\circ$ (mean $80^\circ$). Deviation of the trachea toward a hemithorax may indicate ipsilateral volume loss (e.g., lobar collapse or fibrosis), while contralateral displacement suggests a space-occupying process such as a large pleural effusion or tension pneumothorax. Intraluminal masses, extrinsic narrowing, and splaying of the carina are additional findings sought at this stage.

\paragraph{B~---~Breathing (Lungs, Pulmonary Vessels, and Pleural Spaces).}
Both lungs should be symmetrically expanded and of similar volume. The lung parenchyma is assessed for focal consolidation, ground-glass opacification, interstitial changes, nodules, and masses. Pulmonary vascularity is examined for redistribution (upper lobe diversion) or congestion. The pleural spaces are inspected for effusion (blunting of the costophrenic angles), pneumothorax (visible pleural line with absent lung markings peripherally), or pleural thickening.

\paragraph{C~---~Circulation (Cardiomediastinal Contour and Great Vessels).}
The cardiac silhouette is assessed for size and morphology. On a standard posteroanterior (PA) film, the cardiothoracic ratio should not exceed 0.5; values above this threshold indicate cardiomegaly. Individual cardiac borders are inspected for clarity, as loss of definition localises underlying parenchymal pathology (e.g., right heart border effacement in right middle lobe consolidation). The mediastinal contours are reviewed for widening, masses, and lymphadenopathy. The aortic knuckle, aortopulmonary window, and hilar structures are specifically assessed, along with the pulmonary vasculature for signs of hypertension or oedema.

\paragraph{D~---~Diaphragm and Below.}
The right hemidiaphragm normally sits 1--2~cm higher than the left. Both domes should have smooth contours with sharp, acute costophrenic angles bilaterally. Elevation of a hemidiaphragm may indicate phrenic nerve palsy, lobar collapse, or a subphrenic process. Free subdiaphragmatic air (pneumoperitoneum) is a critical finding indicative of visceral perforation and is seen as a radiolucent crescent beneath the diaphragm on erect films. Subdiaphragmatic structures visible on the CXR---including gastric distension, hepatosplenomegaly, and abdominal calculi---are also reviewed at this stage.

\paragraph{E~---~External (Chest Wall, Bones, and Soft Tissues).}
The osseous structures are examined systematically: the clavicles, scapulae, ribs (posterior and anterior arches), sternum, and visible thoracic vertebrae are assessed for fractures, lytic lesions, and deformities. Rib fractures are of particular clinical significance as they may indicate an underlying pulmonary contusion or haemopneumothorax. Soft tissues are reviewed for subcutaneous emphysema, asymmetric breast shadows, and abnormal calcifications or masses.

\paragraph{F~---~Foreign Material (Devices, Lines, and Tubes).}
The final step involves cataloguing all implanted or inserted devices and verifying correct positioning. Endotracheal tube tips should lie 3--5~cm above the carina. Nasogastric tubes should course below the diaphragm into the gastric body. Central venous catheters should terminate at the cavoatrial junction. Pacemaker and implantable cardioverter-defibrillator (ICD) leads are traced from the generator (typically in the left infraclavicular region) to their respective cardiac chambers. Surgical clips, prosthetic heart valves, and other foreign bodies are also documented.

\begin{table}
  \caption{Summary of the ABCDEF systematic chest radiograph review protocol.}
  \label{tab:abcdef}
  \centering
  \renewcommand{\arraystretch}{1.35}
  \begin{tabular}{p{1.6cm} p{2.4cm} p{8.5cm}}
    \toprule
    \textbf{Letter} & \textbf{Category} & \textbf{Key Findings to Assess} \\
    \midrule
    \textbf{A} & Airways
      & Tracheal position and deviation; carinal angle ($<105^\circ$); main bronchi; intraluminal masses or narrowing \\[4pt]
    \textbf{B} & Breathing
      & Lung parenchyma (consolidation, nodules, interstitial changes); pulmonary vascularity; pleural effusion; pneumothorax \\[4pt]
    \textbf{C} & Circulation
      & Cardiothoracic ratio ($<0.5$ on PA); cardiac borders; mediastinal width and contour; aortic knuckle; hilar structures \\[4pt]
    \textbf{D} & Diaphragm
      & Hemidiaphragm height; costophrenic angles; pneumoperitoneum; subdiaphragmatic pathology (hepatosplenomegaly, calculi) \\[4pt]
    \textbf{E} & External
      & Ribs, clavicles, scapulae, vertebrae for fractures or lesions; soft tissues (subcutaneous emphysema, masses) \\[4pt]
    \textbf{F} & Foreign Material
      & Endotracheal tubes; central venous catheters; nasogastric tubes; pacemaker/ICD leads; prostheses; surgical clips \\
    \bottomrule
  \end{tabular}
\end{table}


\section{Overview of MedPAO Components}
\label{app:medpao}
The MedPAO framework\cite{medpao} introduced an LLM-driven agent designed to transform unstructured, free-text Chest X-Ray (CXR) findings into standardized, protocol-compliant structured reports. Coordinated by a central LLM orchestrator, the system utilizes a suite of clinically validated tools:

\begin{enumerate}
    \item \textbf{Concept Extractor:} This tool distills raw clinical text from radiologists into discrete key findings. By shifting the architecture to a concept-centric paradigm, the system significantly reduces the noise and redundancy inherent in natural language. Powered by a medical LLM fine-tuned on a mixture of expert-annotated and synthetic corpora, this tool mitigates hallucinations by outputting both the discrete concepts and their corresponding source sentences. These sentence pairings subsequently serve as explainable reasoning traces for clinical auditing.
    
    \item \textbf{Ontology Mapper:} This module grounds the extracted concepts within standardized medical ontologies, specifically SNOMED CT. Operating via an external API call, it queries the ontology to retrieve a candidate set of matching terms, along with their associated descriptions and hierarchical ancestors, for each extracted clinical finding.
    
    \item \textbf{Ontology Filter:} Due to the one-to-many nature of ontological mapping, a reasoning model is employed to disambiguate the candidate terms. Utilizing the contextual evidence provided by the source sentences from the Concept Extractor, this tool classifies the retrieved SNOMED CT candidates into primary and secondary matches for each concept.
    
    \item \textbf{Concept Categorizer:} This reasoning module maps the filtered, grounded concepts to appropriate target categories defined by the specified clinical protocol (e.g., the ABCDEF format).
    
    \item \textbf{Structured Report Generator:} A general-purpose LLM synthesizes the final output. Rather than merely concatenating the extracted findings, it leverages the categorized concepts, their source sentences, and the structural guidelines of the clinical protocol to generate a cohesive, medically accurate, and naturally phrased structured report.
\end{enumerate}

\begin{figure}[h]
    \centering
    \includegraphics[width=1\linewidth]{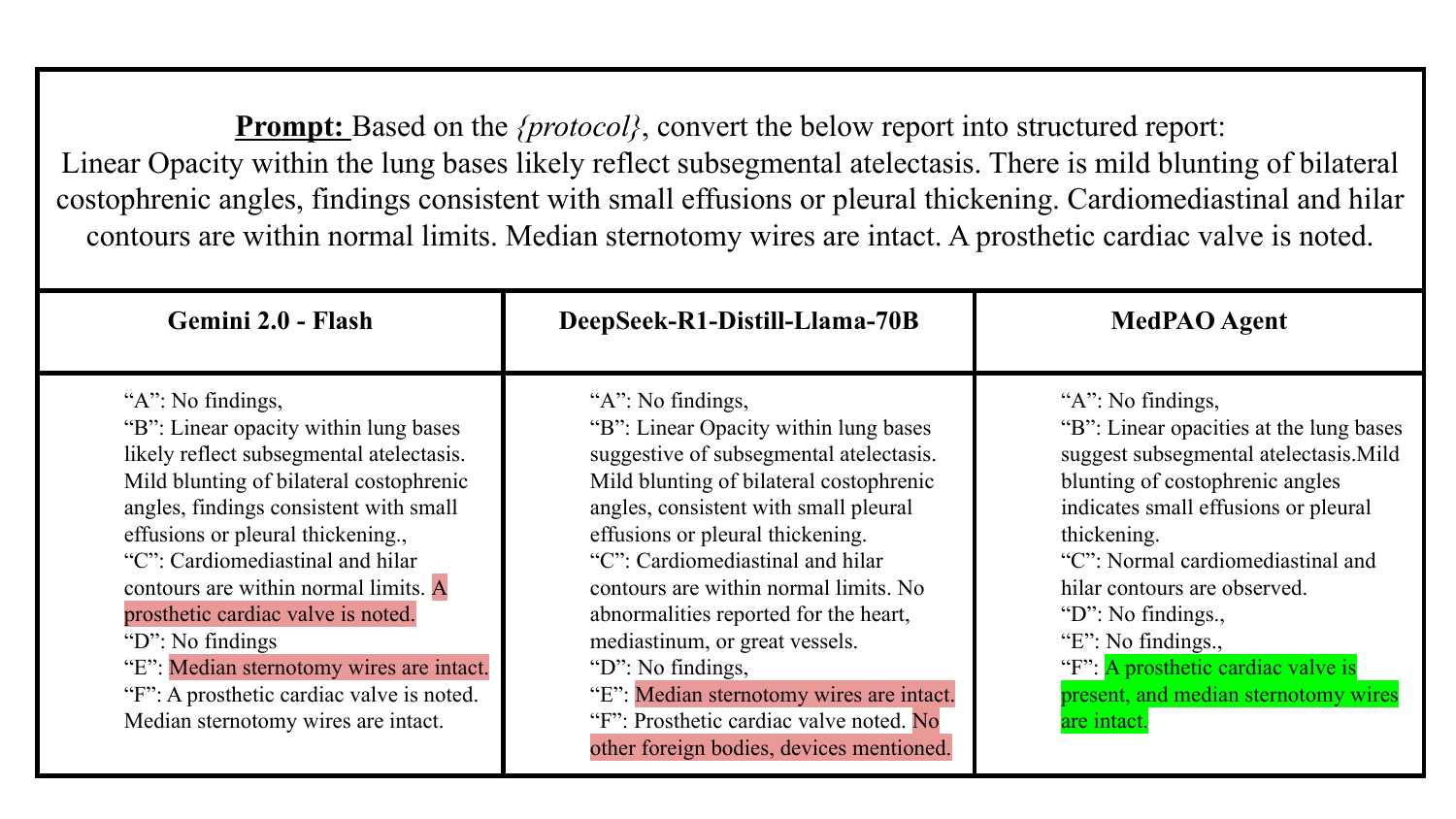}
    \caption{Illustration of MedPAO Agent performance against other models on a hypothetical report. (In the above, the \textit{{protocol}} placeholder represents the injected protocol-related instructions.)}
    \label{fig:agent_vs_llm}
\end{figure}

The proposed MedPAO-VL architecture inherits and integrates several of these core tools for ontology grounding, concept extraction, and clinical NLP tasks. Furthermore, Fig.~\ref{fig:agent_vs_llm} illustrates the comparative effectiveness of the MedPAO Agent against general-purpose LLMs on a representative CXR report. 


\section{Implementation Details}
\label{app:training}
\subsection{Plan-Act-Observe: Technical Implementation}
\label{app:pao_details}

\subsubsection{Semantic Matching Logic}
The overlap checker utilizes the SapBERT\cite{sapbert} model to embed both the vision-module labels
and the concepts extracted from generated free-style reports. We utilize the embedding based cosine similarity metric to assess the overlapping or semantically matching concepts. It follows the algorithm shown in Algorithm \ref{alg:missed_concepts}.

\begin{algorithm}[t]
\caption{Identification of Missed Concepts using Semantic Similarity}
\label{alg:missed_concepts}
\begin{algorithmic}

\Require Vision concepts $V$, Freestyle concepts $F$, threshold $\tau$
\Ensure Missed concepts list $M$

\State $M \gets \emptyset$

\For{each $v \in V$}

    \If{$\text{predicted}(v) = \text{Absent}$}
        \State $v \gets$ ``no '' $\|$ $v$
    \EndIf

    \State $e_v \gets \text{SapBERT}(v)$
    \State $j \gets 0$

    \For{each $f \in F$}

        \State $e_f \gets \text{SapBERT}(f)$

        \If{$\text{cosine\_similarity}(e_v, e_f) > \tau$}
            \State \textbf{break}
        \EndIf

        \State $j \gets j + 1$

    \EndFor

    \If{$j = |F|$}
        \State $M \gets M \cup \{v\}$
    \EndIf

\EndFor

\State \Return $M$

\end{algorithmic}
\end{algorithm}

\subsubsection{Self-Correction Prompting Strategy}
After performing the overlap verification, if the set of missing concepts is non-empty, the \textit{Orchestrator} receives a signal from the \textit{Observe} state indicating that narratives corresponding to the uncovered concepts must be generated. The orchestrator then prompts the concept-grounded report generator with the CXR image along with the list of missing concepts, instructing it to generate narrative descriptions only for those missing findings.

Following this extended report generation step, concepts are again extracted from the updated report and the overlap checker is executed to verify whether all vision-derived concepts are now present in the generated narrative. If some concepts remain uncovered, the procedure is repeated for a predefined number of iterations $N$. Conversely, once the generated freestyle report successfully incorporates all concepts predicted by the vision module, the iterative loop terminates and the orchestrator proceeds to the subsequent protocol-based report generation stage.


\subsubsection{Concept Cache Architecture and Persistence}
The Concept Cache is implemented as a persistent Key-Value store. Keys are the raw concepts the system has encountered previously, and values are standardized ABCDEF category assignments. The process of concept categorization is the same as proposed in \cite{medpao}, which has been validated on a radiologist annotated gold standard dataset. The Concept cache is updated after each successful execution of the agent(i.e., after generation of each structured report). 

Table~\ref{tab:concept_cache_examples} lists a subset of frequently observed concepts stored in the Concept Cache along with their corresponding ABCDEF category assignments. These concepts represent commonly occurring
findings and normal observations encountered during chest X-ray report generation.

\begin{table}[h]
\centering
\caption{Examples of frequently occurring concepts stored in the Concept Cache.}
\label{tab:concept_cache_examples}
\begin{tabular}{ll}
\hline
\textbf{Concept} & \textbf{Category} \\
\hline
Well inflated lungs & B \\
No focal airspace consolidation & B \\
No pleural effusions & B \\
No pulmonary edema & B \\
No pneumothorax & B \\
Normal cardiac contours & C \\
Normal mediastinal contours & C \\
Normal pulmonary vasculature & B \\
No acute bony abnormality & E \\
No acute osseous abnormalities & E \\
Left-sided pacemaker device & F \\
Clips in right upper quadrant of abdomen & F \\
Unchanged eventration of right hemidiaphragm & D \\
Normal mediastinal and hilar contours & C \\
Normal hilar contours & C \\
\hline
\end{tabular}
\end{table}

\subsection{Concept Grounded Report Generator Implementation}

We used a finetuned Medgemma-4b\cite{medgem_vlm1} Multimodal model for the Concept Grounded Report Generation Tasks. To reduce memory overhead, we loaded the Medgemma-4b\cite{medgem_vlm1} in 4-bit NormalFloat (NF4) quantization. We applied Parameter-Efficient Fine-Tuning (PEFT) via Low-Rank Adaptation (LoRA)~\cite{lora}, with a rank(r) of 32 and an alpha of 64, and a dropout probability of 0.05. Fine-tuning was performed on a 50,000-report subset for four epochs using AdamW as optimizer with a learning rate of $2\times10^{-4}$, a cosine schedule, a batch size of 4, and gradient accumulation over 64 steps. During inference, reports were generated with a maximum of 150 new tokens. All the experiments are done with a random seed of 42. This quantized PEFT setup reduces memory consumption sufficiently to fine-tune on a single NVIDIA A6000 (48 GB) GPU. 


\section{Radiologist Evaluation}
\label{app:rad_metrics}

\subsection{Structured Report Evaluation Metrics}
To assess the quality of the generated structured radiology reports, we introduced two key scoring metrics: (1) an Accuracy Score, which measures the clinical correctness of the generated report against the original MIMIC-CXR reference reports, and (2) a Structural Score, which evaluates the degree to which the generated reports adhere to the defined reporting protocol. These metrics provide a comprehensive assessment of both the medical validity and standardized formatting of the outputs. The assessment process involved presenting radiologists with paired samples - our system-generated structured reports, MedGemma-27b \cite{medgem_vlm1} multimodal model generated structured reports alongside their corresponding MIMIC-CXR images and reference reports. This side-by-side comparison enabled direct evaluation of report quality against established clinical standards. Radiologists evaluated each report pair blindly (without knowing which report corresponds to which method) using the standardized metrics outlined in Tables \ref{tab5} and \ref{tab6}.

\begin{table*}
\centering
\renewcommand{\arraystretch}{1.3} 
\caption{Scoring criteria for report quality}
\label{tab5}
\begin{tabular}{p{1cm} p{10cm}}
\toprule
\textbf{Score} & \textbf{Definition} \\
\midrule
5 &  Perfect Match: The Generated Report is a perfect and complete reflection of the Groundtruth Report. All key details are present, and there is no missing, extraneous, or fabricated (hallucinated) information.
\\
4 & Generally Accurate with Minor Omissions: The Generated Report is highly accurate and contains all key details. It may be missing a few minor, non-critical details that do not impact the overall meaning or interpretation. No hallucinations.
\\
3 & Key Details Present but Incomplete or with Minor Hallucinations: The Generated Report captures the main points but has significant omissions of secondary details. Alternatively, it might contain minor fabricated details that do not fundamentally contradict the groundtruth but add noise. Requires some interpretation to align with the groundtruth.\\
2 & Missing Key Details or Contradictory Information: The Generated Report fails to mention one or more key details from the groundtruth. Alternatively, it contains information that directly contradicts the groundtruth. This report could be misleading but is not immediately dangerous.\\
1 & Critically Flawed or Dangerous: The Generated Report has major omissions of critical information or contains significant, dangerous hallucinations that could lead to incorrect and harmful decisions.\\
\bottomrule
\end{tabular}
\end{table*}

\begin{table*}
\centering
\renewcommand{\arraystretch}{1.3} 
\caption{Scoring criteria for report structure}
\label{tab6}
\begin{tabular}{p{1cm} p{10cm}}
\toprule
\textbf{Score} & \textbf{Definition} \\
\midrule
5 & Perfect Structure: The report is perfectly structured. All specified Key Concepts are present as distinct sections/headers, in the correct order, and contain the relevant information as seen in the groundtruth.\\
4 & Good Structure: All Key Concepts are present and correctly populated, but there might be minor deviations. \\
3 &  Acceptable Structure: Most Key Concepts are present, but one or two may be missing or merged. The information is generally in the right place, with some minor deviations.\\
2 & Poor Structure: Multiple Key Concepts are missing. Information is disorganized, placed under incorrect headings, making the report difficult to follow and non-compliant with the protocol.\\
1 &  No Structure: The report is a free-form block of text. It completely ignores the Key Concepts and shows no attempt to follow the required protocol-driven structure.\\
\bottomrule
\end{tabular}
\end{table*}

\subsection{Extended Radiologist score evaluation}
\begin{table}[h]
\caption{Extended inter-rater statistics for structured 
report evaluation}
\centering
\begin{tabular}{lcccc}
\toprule
Metric & \multicolumn{2}{c}{MedGemma-27B} 
       & \multicolumn{2}{c}{MedPAO-VL} \\
\cmidrule(lr){2-3}\cmidrule(lr){4-5}
 & Accuracy & Structure & Accuracy & Structure \\
\midrule
Combined Mean & $3.18 \pm 0.36$ & $3.98 \pm 0.24$ 
                       & $\mathbf{3.50 \pm 0.28}$ 
                       & $\mathbf{4.28 \pm 0.20}$ \\
\midrule
Cohen's $\kappa$  & 0.510 & 0.394 & 0.222 & 0.369 \\
\bottomrule
\end{tabular}
\end{table}

\begin{figure}
    \centering
    \includegraphics[width=1\linewidth]{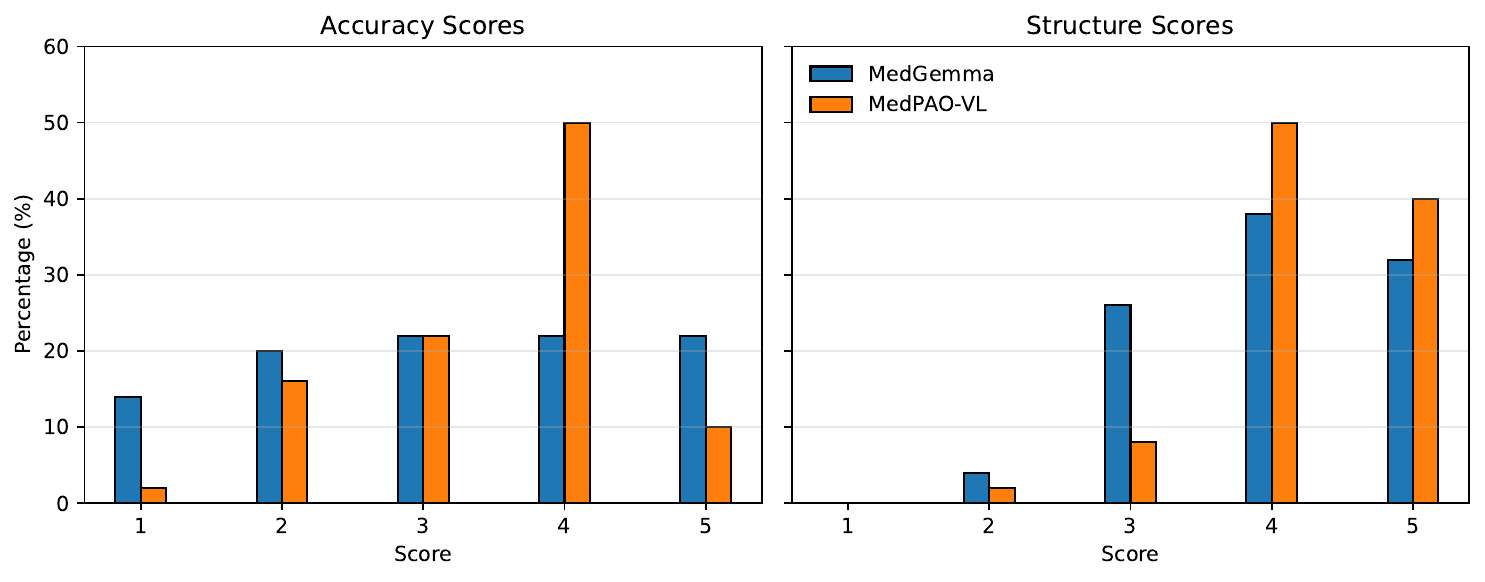}
    \caption{Radiologist score distribution}
    \label{fig:score_distrib}
\end{figure}

\paragraph{Inter-rater Agreement.}
Inter-rater agreement ranged from fair to moderate ($\kappa$ = 0.22--0.51, quadratic weighted), consistent with published radiologist agreement rates on subjective report quality assessment. Tanno et al.~\cite{flamingocxr} highlighted this inherent subjectivity by demonstrating that a panel of 27 radiologists reached unanimous agreement on report preference in only 27.4\% of cases for complex datasets, reflecting the nuanced nature of clinical quality judgments. The lower $\kappa$ observed for MedPAO-VL accuracy (0.222) compared to MedGemma-27B (0.510) is consistent with this phenomenon: outputs of greater clinical nuance are harder to score consistently than reports with clearly identifiable deficiencies. Score distribution analysis (Fig~\ref{fig:score_distrib}) provides a complementary view: 60\% of MedPAO-VL accuracy assessments received a score of 4 or 5, compared to 44\% for MedGemma-27B, and 90\% of MedPAO-VL structure scores reached this threshold versus 70\% for MedGemma-27B. Due to the evaluation cohort size ($N$=50 per rater), formal significance testing is limited in power; a larger multi-reader study is planned as future work.


\section{Data Curation pipeline and Analysis}
\label{app:data_preprocess}

\subsection{Data Curation}
\label{app:data_curation}
\begin{figure}
    \centering
    \includegraphics[width=\textwidth]{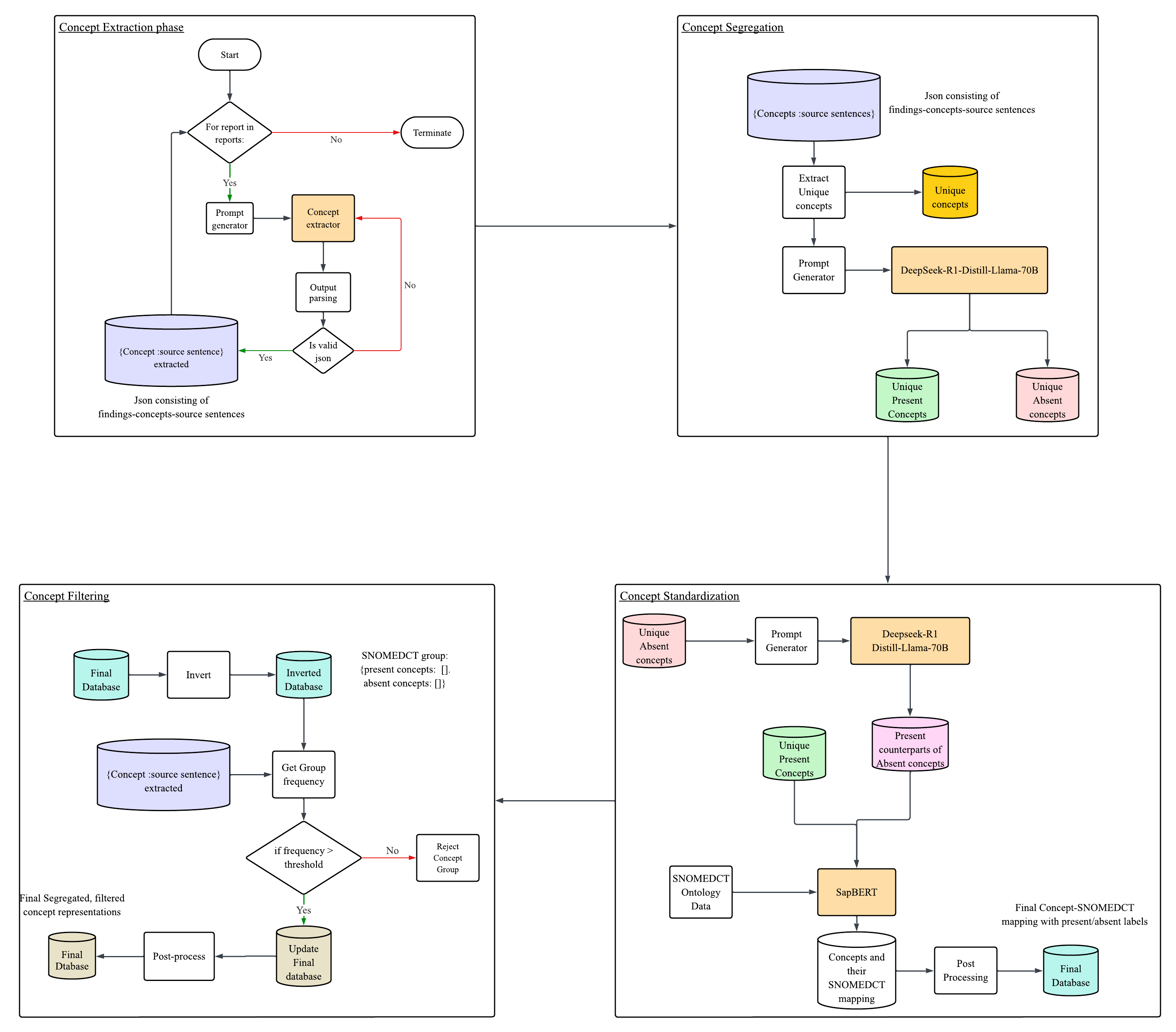}
    \caption{The entire Data preprocessing pipeline for extracting standardized concepts from clinical reports}
    \label{fig:data_preprocess}
\end{figure}

The data curation pipeline comprises four key stages, as illustrated in Figure \ref{fig:data_preprocess}.

\begin{enumerate}
\item \textbf{Concept Extraction:} We leverage the MedPAO concept extractor \cite{medpao} to automatically identify clinical findings within radiology reports. A few-shot prompting strategy is used to query the model for candidate concepts along with the corresponding source sentences, thereby minimizing hallucinations and providing inherent explainability by linking each concept to its originating text span. If the extractor produces invalid or incomplete outputs, the model is re-prompted until a valid set of concepts is obtained. This procedure is applied to all reports, resulting in a structured collection of report–concept–evidence triplets.

\item \textbf{Concept Segregation:} Once all concepts are extracted for each report, the next step involves grouping them into \textit{Present} and \textit{Absent} categories to support downstream standardization. We begin by aggregating the full set of unique extracted concepts (shown in row~1 of Table~\ref{tab:statistic_num_concepts}) and pass them to a large language model with instructions to assign each concept or pathology to one of the two categories. For this classification step, we utilize the DeepSeek-R1-Distill-Llama-70B model~\cite{DeepSeekR1}. This yields two structured outputs: a list of concepts classified as \textit{Present} and another comprising those inferred as \textit{Absent}. The counts of unique \textit{Present} and \textit{Absent} concepts are provided in rows~2 and~3 of Table~\ref{tab:statistic_num_concepts}.

\item \textbf{Concept Standardization:} To enable consistent normalization, we first convert all \textit{Absent} concepts into their corresponding \textit{Present} forms using an LLM (e.g., transforming “no pneumothorax’’ into “pneumothorax’’). These generated \textit{Present counterparts}, together with the originally extracted \textit{Present} concepts, form the unified set used for standardization. For ontology mapping, we employ the state-of-the-art concept normalization model SapBERT \cite{sapbert}, using SNOMED CT as the target ontology due to its comprehensive coverage of clinical findings. SapBERT embeds each concept and matches it to the closest SNOMED CT entry (e.g., CONCEPT ID, ANCESTORS) based on embedding similarity, producing a standardized medical term. Finally, we post-process the outputs by restoring the standardized \textit{Present counterparts} back to their original \textit{Absent} forms. This yields a complete set of concepts—each labeled as \textit{Present} or \textit{Absent} and linked to a corresponding SNOMED CT identifier.

\item \textbf{Concept Filtering:} After obtaining the SNOMED CT mappings for each concept along with their \textit{Present}/\textit{Absent} labels, we construct an inverted index in which each SNOMED CT identifier serves as a key, and its value is a dictionary grouping the corresponding \textit{Present} and \textit{Absent} concepts. This restructuring yields SNOMED CT–centric concept groups (the unique number of such groups is mentioned in row 4 of Table \ref{tab:statistic_num_concepts}), each representing all variants mapped to the same ontology term. Using the original extracted concepts, we compute the occurrence frequency of each SNOMED CT group across the dataset. These frequencies are then used to filter out low-support groups based on a predefined threshold. Specifically, we discard any group with a total frequency below 100. Applying this criterion reduces the overall set of groups to 236, which serve as the final classification labels for our vision model, which can also be seen in row 5 of Table \ref{tab:statistic_num_concepts}.
\end{enumerate}

\begin{table}
    \centering
\caption{Number of concepts in each categories}
\label{tab:statistic_num_concepts}
\begin{tabular}{llcccccc}
        \toprule
        Categories & Number \\
        \midrule
        Unique raw concepts & 60,531  \\
        Present raw concepts & 57,410  \\
        Absent raw concepts & 3,123 \\
        Unique SNOMEDCT groups before filtering & 9,450\\
        Unique SNOMEDCT groups after filtering & 236\\
        \bottomrule
        \end{tabular}
\end{table}

\subsection{Radiologist validation on frequency filtering}
\label{app:radiologist_freq}

Here is the validation given by the radiologist after observing the filtered and rejected concept groups:

\begin{radiologistcomment}
The 236 accepted concept groups represent clinically meaningful findings commonly encountered in routine practice. The rejected concepts were predominantly technical descriptors (e.g., view type, patient positioning, redundant labels) or extremely low-frequency entities with insufficient samples for reliable learning. Excluding these low-frequency and non-pathologic concepts does not materially affect clinical usability. Their removal reduces class imbalance and noise, improving model robustness and reliability. Rare entities can be incorporated in future iterations as more data become available.
\end{radiologistcomment}

\subsection{Data Analysis}
\label{app:data_analysis}
The Data Analysis part mainly includes the below 3 categories useful for the eventual data splitting and training the models.

\begin{figure*}[h!]
    \centering

    \begin{subfigure}{0.48\linewidth}
        \centering
        \includegraphics[width=\linewidth]{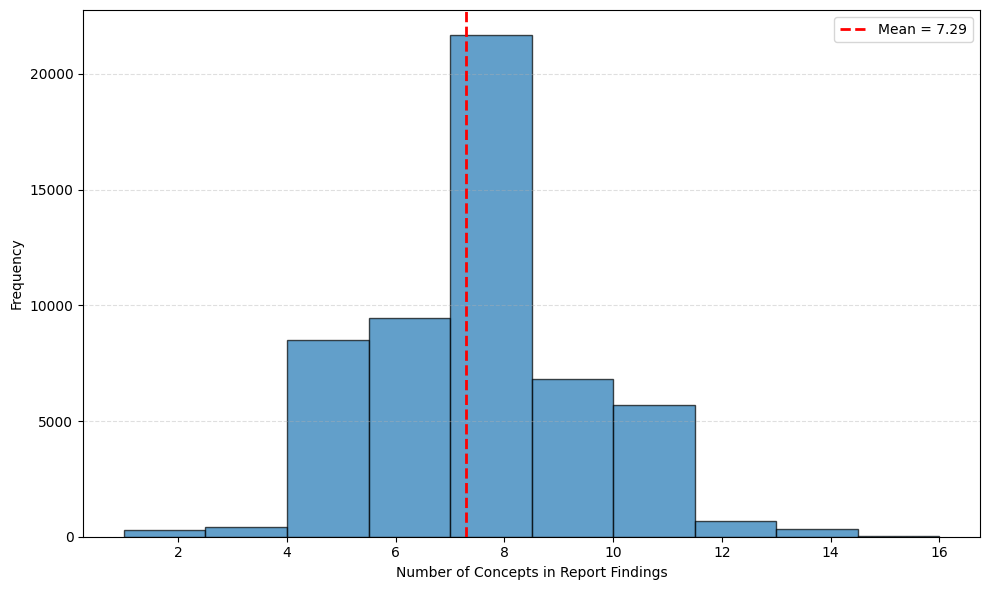}
        \caption{}
        \label{fig:conc_raw}
    \end{subfigure}
    \hfill
    \begin{subfigure}{0.48\linewidth}
        \centering
        \includegraphics[width=\linewidth]{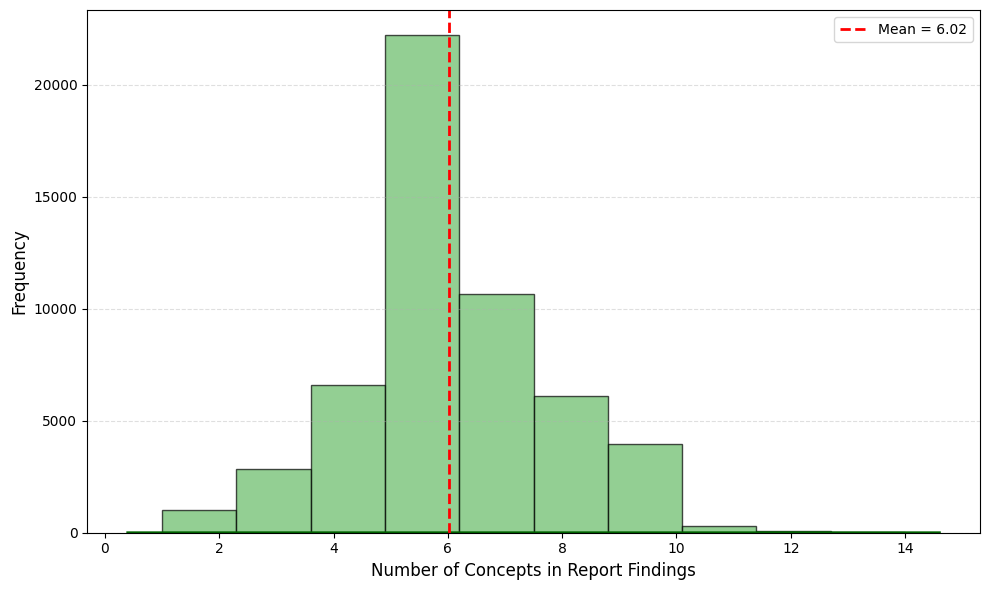}
        \caption{}
        \label{fig:conc_processed}
    \end{subfigure}

    \caption{Comparison of concept distributions across all reports: (a) raw extracted concepts, and (b) standardized concepts after post-processing.}
    \label{fig:conc_per_report}
\end{figure*}

\begin{figure*}[h!]
    \centering
    \includegraphics[width=1\linewidth]{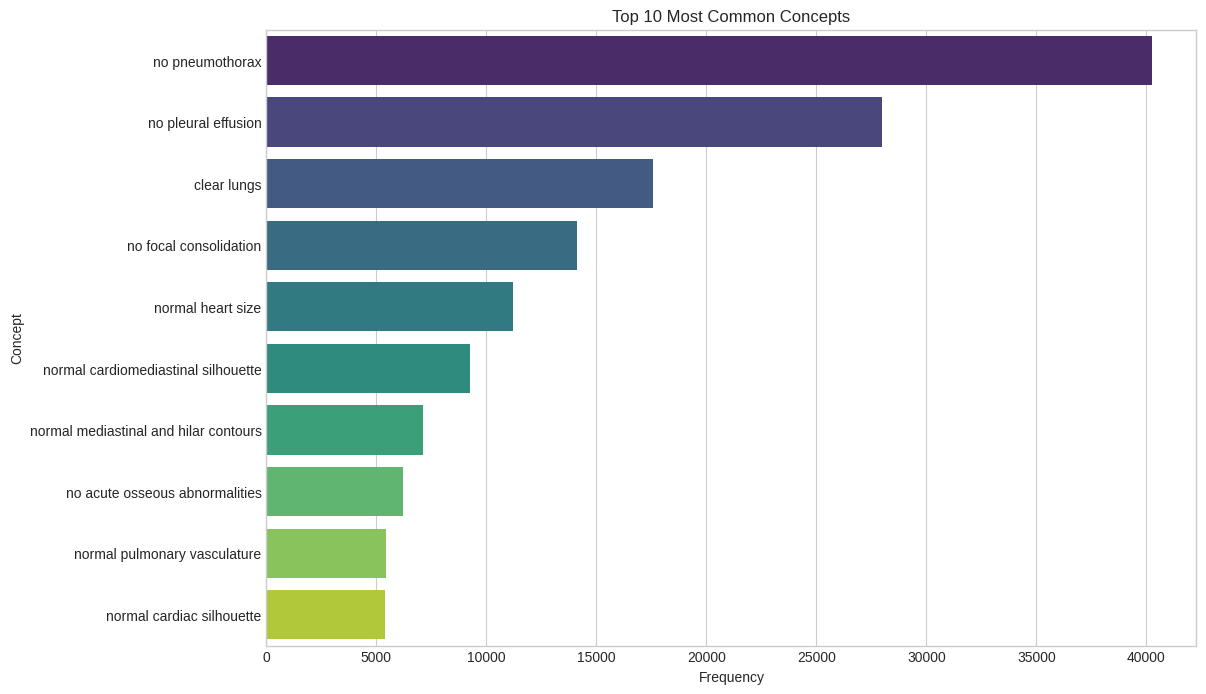}
    \caption{Top 10 raw Concept and their Frequency}
    \label{fig:conc_frequency}
\end{figure*}

\textbf{Concept Distribution per Report:} Figure \ref{fig:conc_per_report} illustrates the distribution of both raw and standardized concepts across all reports in the dataset. As observed, the overall shape of the distributions remains similar; however, the standardized concept distribution (Figure \ref{fig:conc_processed}) exhibits a noticeable leftward shift, reflecting a reduced mean number of concepts per report. This reduction primarily results from the removal of low-frequency concepts during the filtering stage. Also Figure \ref{fig:conc_frequency} represents the Top 10 raw concepts based on their frequency of occurance.

\begin{figure}
    \centering
    \includegraphics[width=1\linewidth]{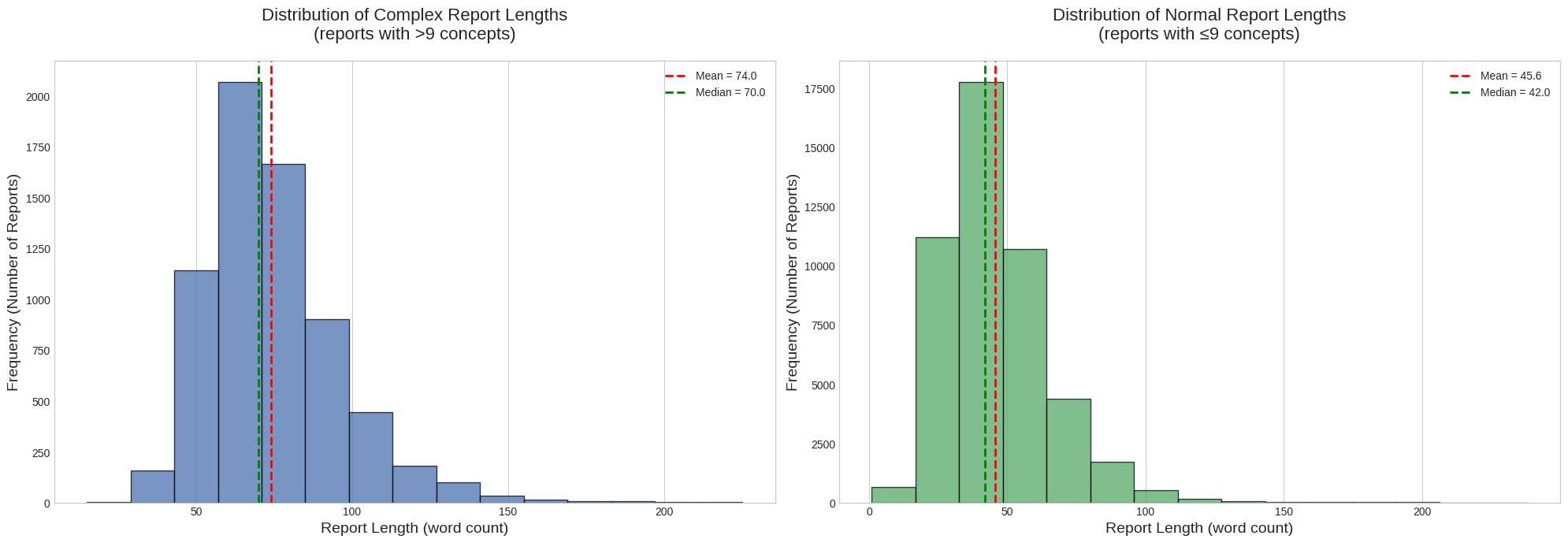}
    \caption{Distribution of no. of concepts across normal and complex reports in the current dataset.}
    \label{fig:normal_complex}
\end{figure}

\textbf{Segregation of Normal and Complex Reports:}  
To facilitate effective training of the language decoder, we partition the dataset into \textit{Normal} and \textit{Complex} reports based on the number of extracted concepts. Normal reports tend to be shorter, containing fewer concepts and fewer words, whereas Complex reports exhibit higher concept density and longer narrative descriptions. Using a threshold of nine concepts, we categorize each report into one of these two groups. Figure \ref{fig:normal_complex} shows the word-count distributions for both categories. As expected, the Complex reports exhibit higher mean and median word counts compared to Normal reports, supporting the validity of our concept-based segmentation strategy. Summary statistics for both groups are provided in Table \ref{tab:statistic_num}. Notably, the dataset contains fewer Complex reports, suggesting that many studies correspond to normal or near-normal findings, or that reporting styles across data sources frequently favor concise descriptions. This stratified separation is instrumental in constructing balanced train-validation-test splits for fine-tuning the grounded report generation model.

\begin{table}
    \centering
\caption{Statistic of Normal and Complex reports}
\label{tab:statistic_num}
        \begin{tabular}{llcccccc}
        \toprule
         & Normal Reports & Complex Reports\\
        \midrule
        Number of reports &  47,176 & 6,721 \\
        Mean (word count) & 45.6 & 74.0 \\
        Median (word count) & 70 & 70.0 \\
        Std Deviation (word count) & 21.5 & 19.0 \\
        1st Quartile (word count) & 32.0 & 59.0 \\
        3rd Quartile (word count) & 56.0 & 85.0 \\
        \bottomrule
        \end{tabular}
\end{table}

\begin{figure}
    \centering
    \includegraphics[width=1\linewidth]{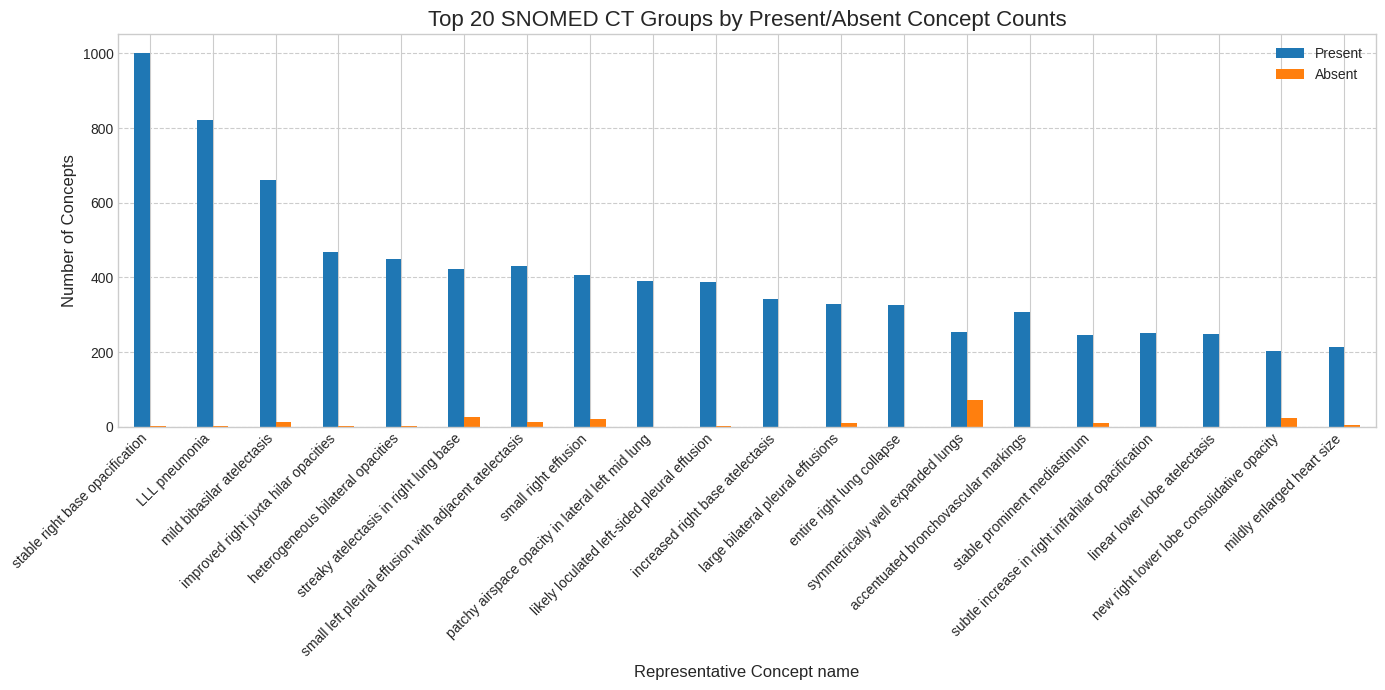}
    \caption{Top SNOMEDCT groups and their Present and Absent concept counts}
    \label{fig:snomedct_concept_counts}
\end{figure}

\begin{figure}
    \centering
    \includegraphics[width=1\linewidth]{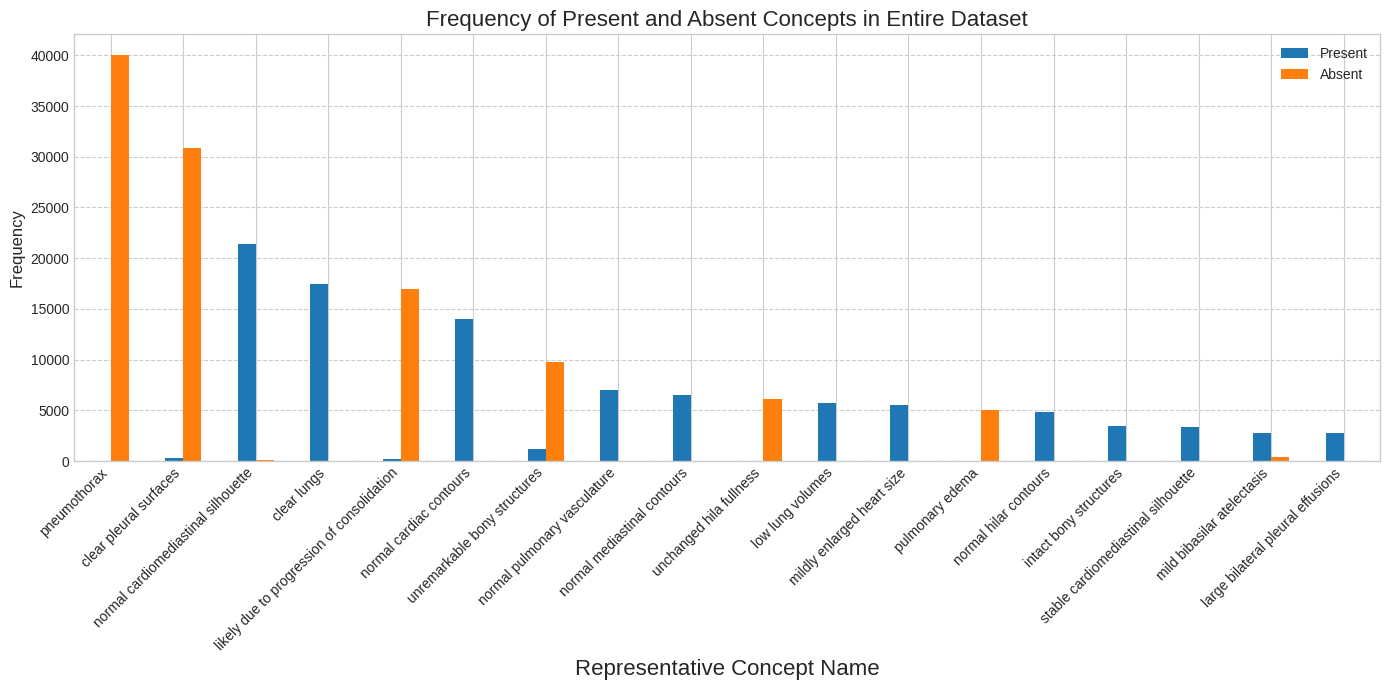}
    \caption{Top SnomedCT concept groups and their Frequency of occurance in the given dataset.}
    \label{fig:snomedct_freq}
\end{figure}

\textbf{Standardized Concept Counts and Frequency Analysis:} 
Figure \ref{fig:snomedct_concept_counts} illustrates the distribution of raw concepts across the SNOMED CT–standardized concept groups. As shown, each group contains a varying number of concepts classified under both the \textit{Present} and \textit{Absent} categories. Following this grouping, a frequency-level analysis was performed to examine how often each concept group appears within the report findings across the dataset. The resulting frequency distribution is depicted in Figure \ref{fig:snomedct_freq}, highlighting the variation in occurrence patterns between the \textit{Present} and \textit{Absent} classes within each standardized concept group.

\section{Extended Qualitative evaluation}
\label{app:extended_evals}

\subsection{Emergent Clinical Co-occurrence in Latent Space}
\label{app:clinical_cooccurance}

\begin{figure*}[h!]
    \centering

    \begin{subfigure}{0.45\linewidth}
        \centering
        \includegraphics[width=\linewidth]{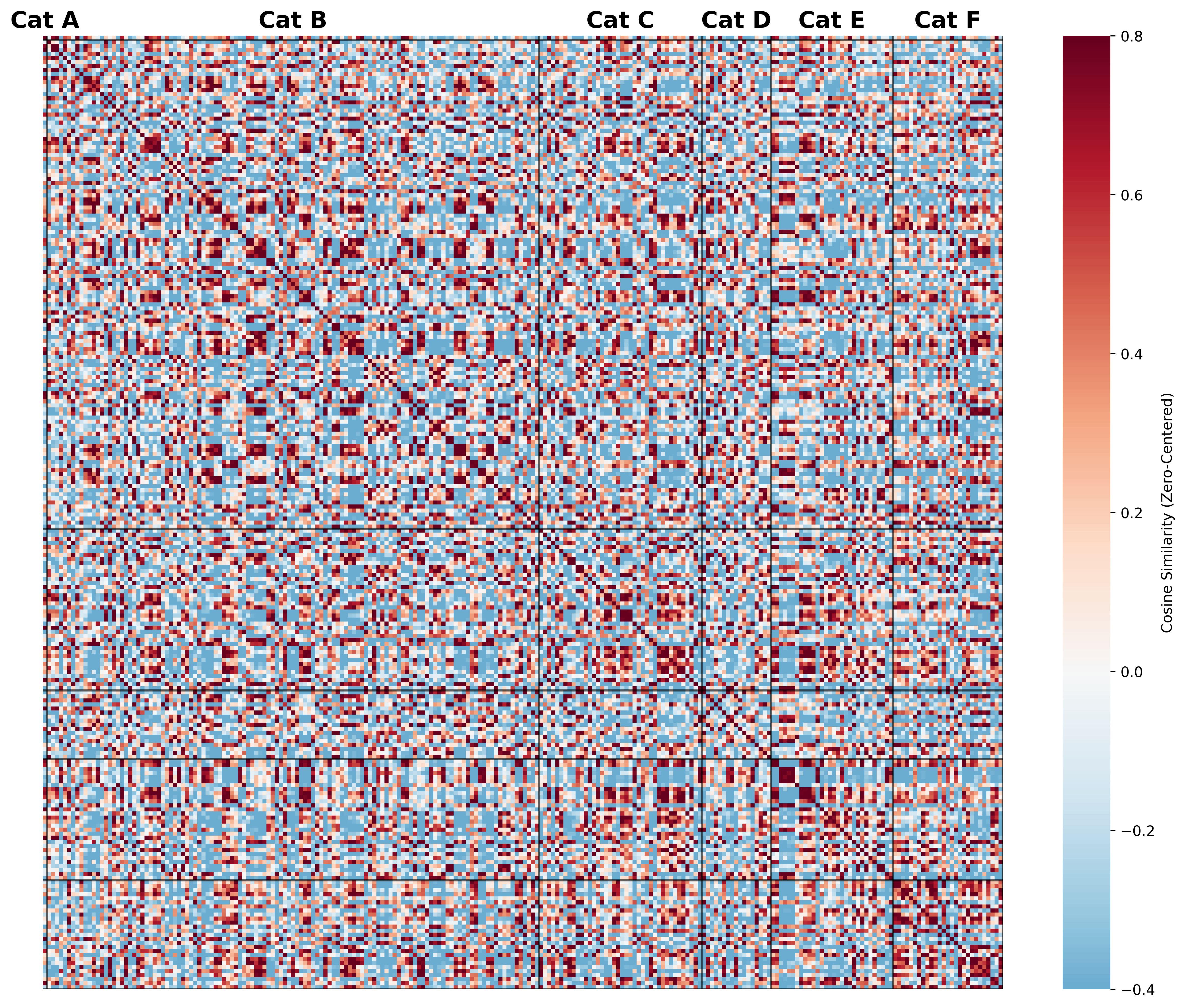}
        \caption{}
        \label{fig:correlation1}
    \end{subfigure}
    \hfill
    \begin{subfigure}{0.45\linewidth}
        \centering
        \includegraphics[width=\linewidth]{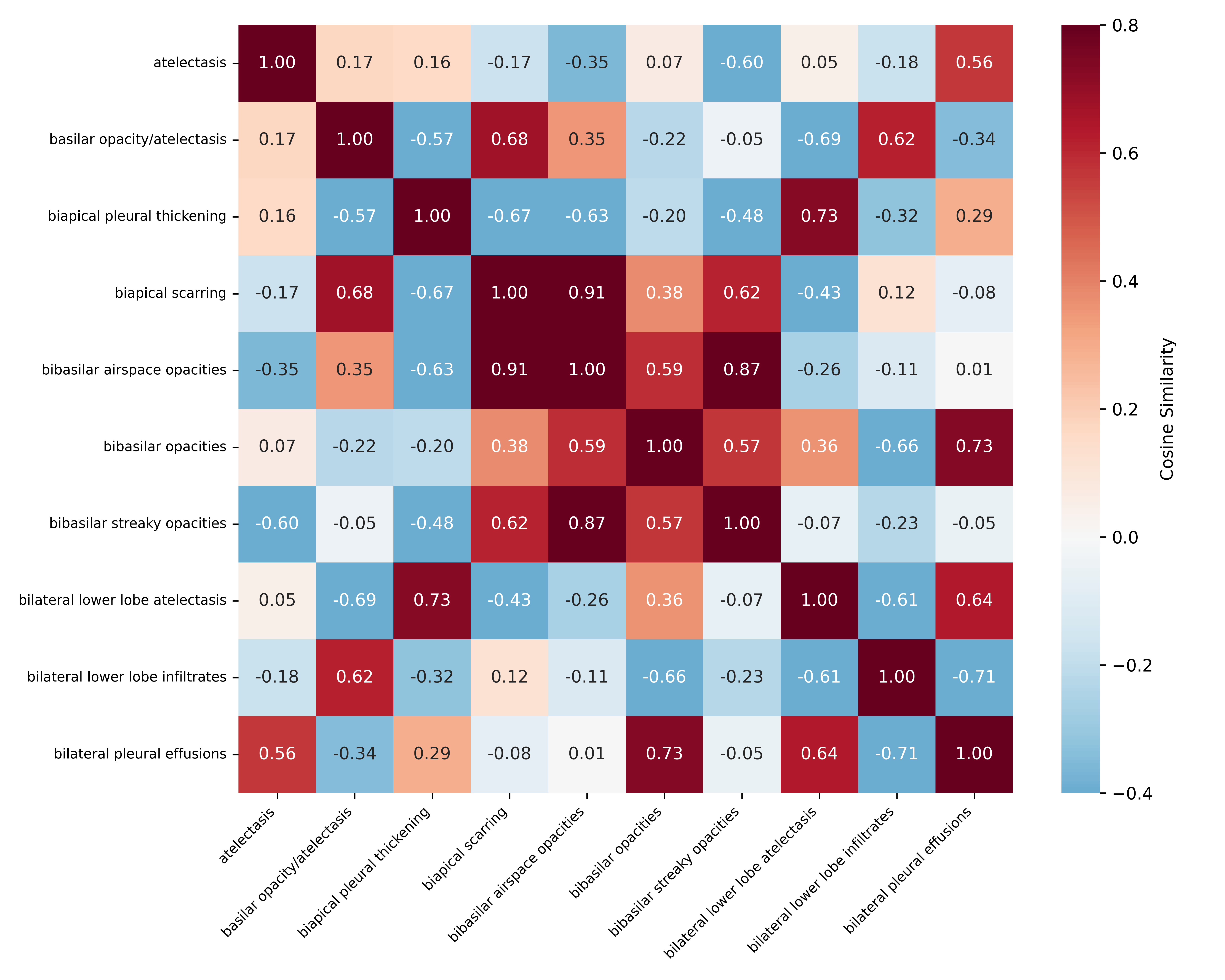}
        \caption{}
        \label{fig:correlation_zoom}
    \end{subfigure}

    \caption{Heatmap showing Cosine similarity amongst the observed pathologies. (a)Cosine similarity amongst all the pathologies. (b)A cross-section of B(Breathing)}
    \label{fig:sim_as_corre}
\end{figure*}

To evaluate whether the learned pathology embeddings of the Pathology Slot Module (PSM) capture real-world clinical relationships, we analyze the cosine similarity matrix of these learned pathology embeddings. As illustrated in the global heatmap encompassing all target concepts (Fig.\ref{fig:correlation1}), the latent space predominantly maintains orthogonality, indicated by the widespread neutral and negative similarities. This is a critical property which confirms that the slots representations successfully resisted feature collapse, thereby preserving the strict diagnostic specificity required to differentiate distinct pathologies.

Amidst this orthogonality, the model exhibits localized blocks of high positive correlation. A high-resolution cross-section of the "Breathing" (Category B) pathologies (Fig. \ref{fig:correlation_zoom}) explicitly validates that these correlations are not noise, but rather emergent encodings of clinical co-occurrence. Rather than isolating every finding independently, the manifold tightly aligns embeddings where clinical presentation frequently overlaps. For instance, \textit{biapical scarring} and \textit{bibasilar airspace opacities} demonstrate a strong semantic alignment with a cosine similarity of 0.91. This effectively mirrors the common co-occurrence of these findings in patients with chronic pulmonary architectural changes, confirming that MedPAO-VL's internal representations natively align with established medical epistemology.

\subsection{Clinical Manifold Analysis and Concept Specialization}
\label{app:manifold_analysis}
\begin{figure}
    \centering
    \includegraphics[width=\textwidth]{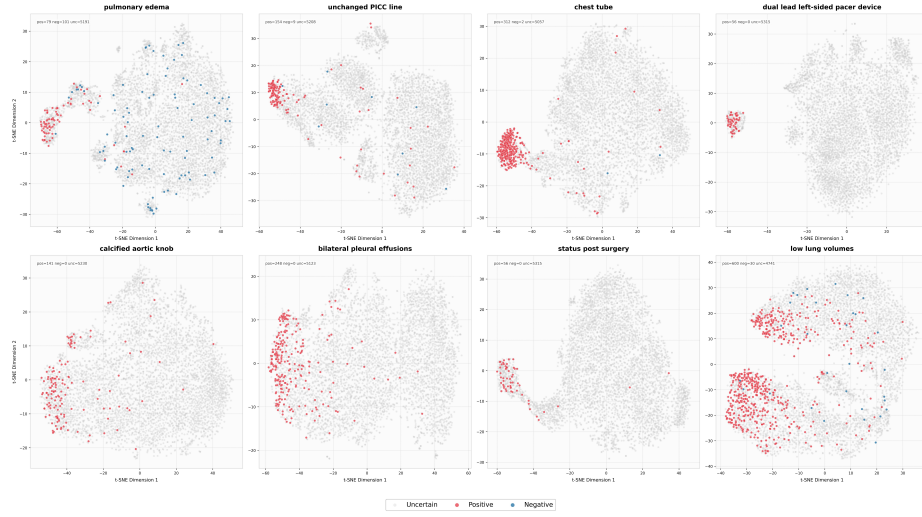}
    \caption{Per-slot t-SNE visualizations for representative clinical findings.}
    \label{fig:clinical_tsne}
\end{figure}

To further investigate the discriminative power of the Pathology Slot Module (PSM), we perform per-pathology t-SNE projections of the context vectors $U_T$ across the test set. A notable observation from figure \ref{fig:clinical_tsne} shows the high degree of latent separability achieved for rare but visually distinct objects, such as \textit{dual lead left-sided pacer devices} and \textit{chest tubes}. Despite the "Uncertain" class dominating the majority of the latent space, the positive clinical instances congregate in highly localized, high-density clusters. Furthermore, for acute pathologies like \textit{pulmonary edema}, the model exhibits a clear class-conditional separation. The positive clusters are spatially removed from the negative background, proving that the PSM maintains diagnostic specificity even in complex, multi-pathological images. These analysis confirms that each slot successfully specializes in a unique visual signature, effectively disentangling medical findings from the global image features. 

\subsection{Impact of Similarity loss on Attention Maps}
\label{app:sim_loss}

\begin{figure}
    \begin{subfigure}{1\textwidth}
        \centering
        \includegraphics[width=\linewidth]{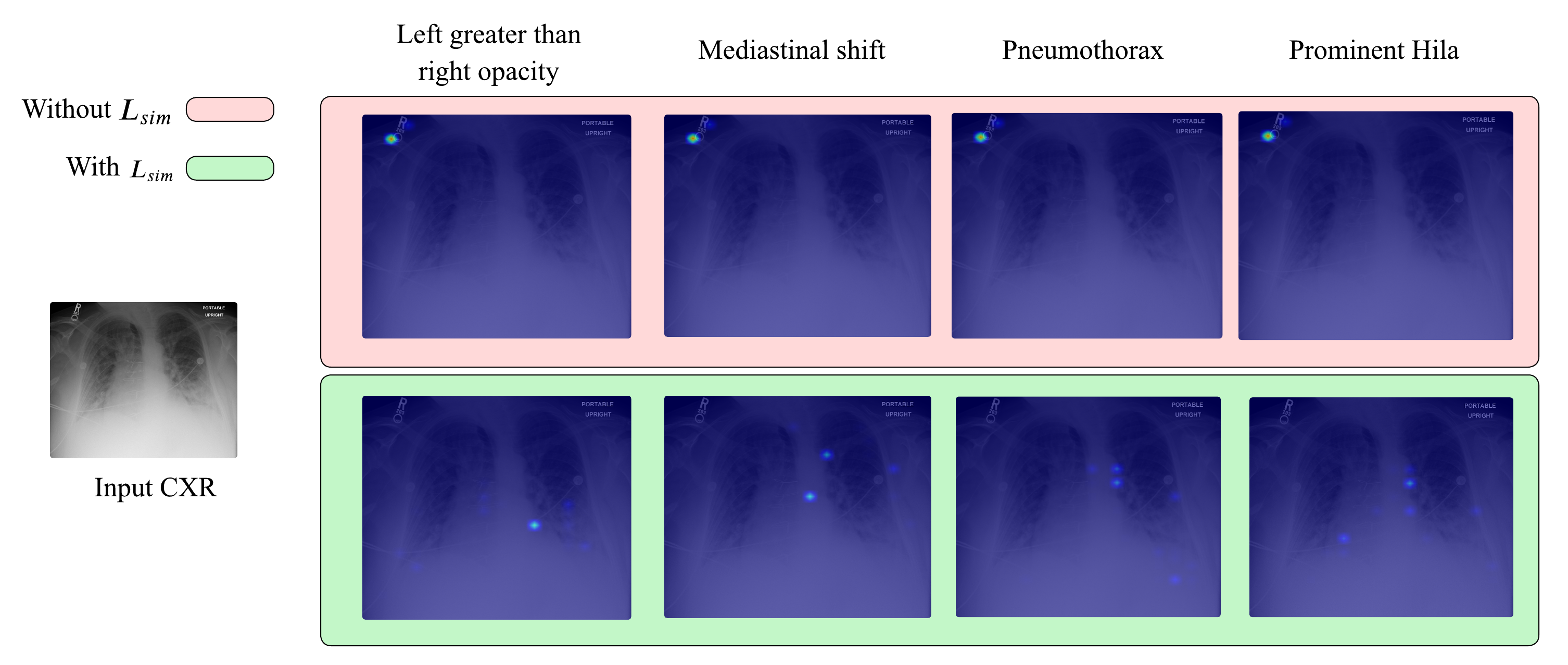}
        \caption{}
        \label{fig:Lsim1}
    \end{subfigure}
    \hfill
    \begin{subfigure}{1\textwidth}
        \centering
        \includegraphics[width=\linewidth]{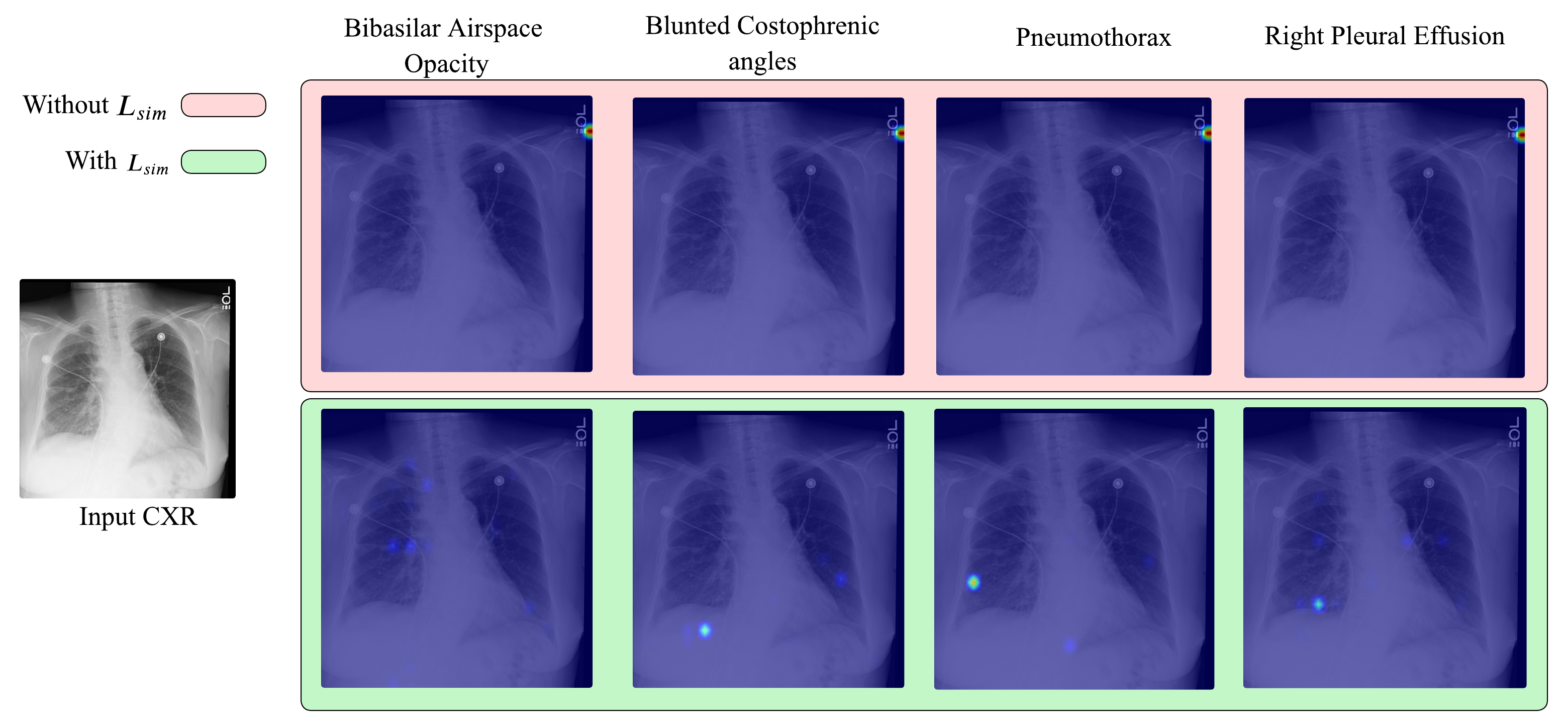}
        \caption{}
        \label{fig:Lsim2}
    \end{subfigure}
    \caption{Attention map comparison between models with and without $L_{sim}$}
    \label{fig:combined_Lsim}
\end{figure}

As discussed Loss Formulation, the initial formulation of the model incorporating the Pathology Slot Module failed to produce spatially distinct attention maps for different pathologies. To mitigate this issue, a similarity loss ($L_{sim}$) was introduced to discourage redundant attention patterns across slots. The qualitative impact of this modification is illustrated in Figure~\ref{fig:combined_Lsim}, which presents a side-by-side comparison of the attention maps generated with and without the inclusion of $L_{sim}$.

In the absence of the similarity loss, the model tends to adopt a ``lazy'' strategy by concentrating its attention on a limited region of the image. This behavior is evident in Figure~\ref{fig:Lsim1}, where the model consistently focuses on the top-left corner irrespective of the pathology being considered. In contrast, incorporating $L_{sim}$ encourages the model to distribute its attention more diversely across relevant regions.

A similar pattern is observed in Figure~\ref{fig:Lsim2}, where the model without the similarity loss predominantly attends to the top-right edge of the image. The introduction of $L_{sim}$ alleviates this issue, resulting in more spatially meaningful attention maps that better correspond to the underlying pathologies. Therefore, beyond improving the quantitative performance of the model, the similarity loss plays a crucial role in enhancing the interpretability and quality of the generated explanation maps.

 In this degenerate regime, the model satisfies the classification 
objective through a ``lazy'' strategy that ignores slot-specific distinctions, 
yielding attention maps that are visually indistinguishable and therefore 
clinically uninformative.


\section{End-to-End System Walkthrough}
\label{app:E2Ewalkthrough}

To demonstrate the auditable reasoning capabilities of MedPAO-VL, we provide a step-by-step trace of the system processing a highly complex case. This example highlights how the agent handles diagnostic uncertainty and self-corrects semantic discrepancies before generating the final protocol-compliant report.

\begin{figure}
    \centering
    \includegraphics[trim=0cm 0cm 0cm 8.1cm, clip,width=1\linewidth]{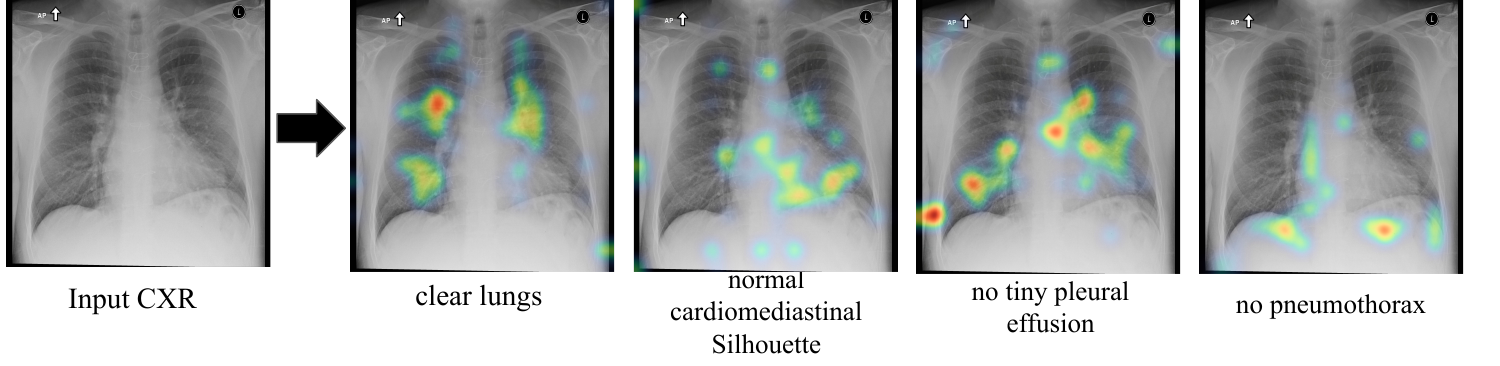}
    \caption{Input CXR and the explainable heatmaps from Vision Concept Module}
    \label{fig:E2E_attn}
\end{figure}

\begin{phaselist}

\item Visual Perception
\begin{modulelist}
    \item \module{Vision Concept Module}
    \begin{iolist}
        \item \inp{Raw CXR Image}
        \item \proc{Extracts visual features, projects to pathology slots, and estimates present/absent/uncertain probabilities per pathology along with spatial attention maps. The attention maps extracted for this particular sample under consideration is shown in Fig~\ref{fig:E2E_attn}}
        \item \outp{\textit{Present:} clear lungs, normal cardiomediastinal silhouette, status post sternotomy and coronary artery bypass graft surgery; \textit{Absent:} tiny pleural effusion, pneumothorax}
    \end{iolist}
\end{modulelist}

\item Initial Report Generation
\begin{modulelist}
    \item \module{Concept-Grounded Report Generator}
    \begin{iolist}
        \item \inp{CXR Image; \textit{Present:} clear lungs, normal cardiomediastinal silhouette, status post sternotomy and coronary artery bypass graft surgery; \textit{Absent:} tiny pleural effusion, pneumothorax}
        \item \proc{Generates initial narrative findings based strictly on the grounded visual concepts and CXR analysis.}
        \item \textit{Output:}
        \begin{codeout}
        FINDINGS: The lungs are well expanded and clear. Cardiomediastinal and hilar contours are stable. Median sternotomy wires and mediastinal clips are noted.
        \end{codeout}
    \end{iolist}
\end{modulelist}

\item Self Correction Loop

\textbf{Iteration 1}
\begin{modulelist}
    \item \module{Concept Extractor}
    \begin{iolist}
        \item \textit{Input:}
        \begin{codeout}
            FINDINGS: The lungs are well expanded and clear. Cardiomediastinal and hilar contours are unremarkable. Median sternotomy wires and mediastinal clips are noted. There is no pleural effusion or pneumothorax.
        \end{codeout}
        \item \proc{Extracts clinical findings and corresponding source sentences from the previously generated report.}
        \item \textit{Output:}
        \pydict{
          clear lungs/The lungs are well expanded and clear.,
          stable cardiomediastinal contours/Cardiomediastinal and hilar contours are stable.,
          stable hilar contours/Cardiomediastinal and hilar contours are stable.,
          sternotomy wires/Median sternotomy wires and mediastinal clips are noted,
          mediastinal clips/Median sternotomy wires and mediastinal clips are noted%
        }
    \end{iolist}

    \item \module{Overlap Checker}
    \begin{iolist}
        \item \textit{Input:} two concept sets to compare — \\
            \textbf{Report concepts:}
            \pylist{clear lungs, stable cardiomediastinal contours, stable hilar contours, sternotomy wires, mediastinal clips}
            \textbf{Vision concepts:}
            \pylist{clear lungs, normal cardiomediastinal silhouette, status post sternotomy and coronary artery bypass graft surgery, no tiny pleural effusion, no pneumothorax}
        \item \proc{Utilizes SapBERT for embedding-based matching between the concepts extracted from the report and the concepts from the vision module, and returns the missing concepts.}
        \item \textit{Output:}
        \pylist{no tiny pleural effusion, no pneumothorax}
    \end{iolist}

    \item \module{Re-prompt Report Generator}
    \begin{iolist}
        \item \textit{Input:} \\
        \textbf{Previous Report:} \\
        \begin{codeout}
        FINDINGS: The lungs are well expanded and clear. Cardiomediastinal and hilar contours are unremarkable. Median sternotomy wires and mediastinal clips are noted.
        \end{codeout}
        \textbf{Missing Concepts:}
        \pylist{no tiny pleural effusion, no pneumothorax} 
        \item \proc{Takes the previously generated report and the set of missing concepts from the vision module, and generates the missing piece of the report.}
        \item \textit{Output:}
        \begin{codeout}
        FINDINGS: The lungs are well expanded and clear. Cardiomediastinal and hilar contours are unremarkable. Median sternotomy wires and mediastinal clips are noted. There is no pleural effusion or pneumothorax.
        \end{codeout}
    \end{iolist}
\end{modulelist}

\vspace{0.5em}
\textbf{Iteration 2}
\begin{modulelist}
    \item \module{Concept Extractor}
    \begin{iolist}
        \item \textit{Input:}
        \begin{codeout}
            FINDINGS: The lungs are well expanded and clear. Cardiomediastinal and hilar contours are unremarkable. Median sternotomy wires and mediastinal clips are noted. There is no pleural effusion or pneumothorax.
        \end{codeout}
        
        \item \proc{Extracts clinical findings and corresponding source sentences from the previously generated report.}
        \item \textit{Output:}
        \pydict{
          clear lungs/The lungs are well expanded and clear.,
          stable cardiomediastinal contours/Cardiomediastinal and hilar contours are stable.,
          stable hilar contours/Cardiomediastinal and hilar contours are stable.,
          sternotomy wires/Median sternotomy wires and mediastinal clips are noted.,
          mediastinal clips/Median sternotomy wires and mediastinal clips are noted.,
          no pleural effusion/There is no pleural effusion or pneumothorax.,
          no pneumothorax/There is no pleural effusion or pneumothorax.
        }
        
    \end{iolist}

    \item \module{Overlap Checker}
    \begin{iolist}
        \item \textit{Input:} two concept sets to compare — \\
            \textbf{Report concepts:}
            \pylist{clear lungs, stable cardiomediastinal contours, stable hilar contours, sternotomy wires, mediastinal clips, no pleural effusion, no pneumothorax}
            \textbf{Vision concepts:}
            \pylist{clear lungs, normal cardiomediastinal silhouette, status post sternotomy and coronary artery bypass graft surgery, no tiny pleural effusion, no pneumothorax}
        \item \proc{Utilizes SapBERT for embedding-based matching between the concepts extracted from the report and the concepts from the vision module, and returns the missing concepts.}
        \item \outp{[] \quad $\rightarrow$ \textbf{Loop exits}}
    \end{iolist}
\end{modulelist}

\item Report Structuring
\begin{modulelist}
    \item \module{Concept Extractor}
    \begin{iolist}
        \item \textit{Input:}
        \begin{codeout}
        FINDINGS: The lungs are well expanded and clear. Cardiomediastinal and hilar contours are unremarkable. Median sternotomy wires and mediastinal clips are noted. There is no pleural effusion or pneumothorax.
        \end{codeout}
        \item \proc{Extracts clinical findings and corresponding source sentences from the previously generated report.}
        \item \textit{Output:}
        \pydict{
          clear lungs/The lungs are well expanded and clear.,
          stable cardiomediastinal contours/Cardiomediastinal and hilar contours are stable.,
          stable hilar contours/Cardiomediastinal and hilar contours are stable.,
          sternotomy wires/Median sternotomy wires and mediastinal clips are noted.,
          mediastinal clips/Median sternotomy wires and mediastinal clips are noted.,
          no pleural effusion/There is no pleural effusion or pneumothorax.,
          no pneumothorax/There is no pleural effusion or pneumothorax.
        }
    \end{iolist}

    \item \module{Ontology Mapper}
            \begin{iolist}
                \item \textit{Input:}
                \pylist{clear lungs, stable cardiomediastinal silhouette, stable hilar contours, sternotomy wires, mediastinal clips, no tiny pleural effusion, no pneumothorax}
            
                \item \proc{Maps each of the extracted concepts to the SNOMED CT ontology to obtain further clinical information for structural grounding in text.}
            
                \item \textit{Output:}
                \begin{codelist}
            [
              {
                "query": "clear lungs",
                "concept_id": "444859001",
                "matched_surface": "Lungs in normal arrangement",
                "preferred_desc": "Lungs in normal arrangement (finding)",
                "all_descs": [
                  "Lungs in normal arrangement (finding)",
                  "Lungs in normal arrangement"
                    ],
                "ancestors": [
                  "Normal lung (finding)",
                  "Normal respiratory system (finding)",
                  "Lung finding (finding)",
                  "Respiratory finding (finding)",
                  "Finding of thoracic region (finding)",
                  "Lower respiratory tract finding (finding)"
                    ]
              },
              {
                "query": "stable cardiomediastinal silhouette",
                "concept_id": "233819005",
                "matched_surface": "Stable angina (disorder)",
                "preferred_desc": "Stable angina (disorder)",
                "all_descs": [
                  "Stable angina (disorder)",
                  "Stable angina"
                    ],
                "ancestors": [
                  "Chronic ischemic heart disease (disorder)",
                  "Angina (disorder)",
                  "Chronic chest pain (finding)",
                  "Ischemic heart disease (disorder)",
                  "Chronic heart disease (disorder)"
                    ]
              },
              {
                "query": "no pneumothorax",
                "concept_id": "49756002",
                "matched_surface": "Perinatal pneumothorax (disorder)",
                "preferred_desc": "Perinatal pneumothorax (disorder)",
                "all_descs": [
                    "Perinatal pneumothorax (disorder)",
                    "Perinatal pneumothorax"
                ],
                "ancestors": [
                    "Pneumothorax (disorder)"
                   "Neonatal respiratory system disorder (disorder)"
                    "Fetal intrapartum and/or early neonatal disorder (disorder)"
                    "Disorder of pleura AND/OR pleural cavity (disorder)"
                    "Neonatal disorder (disorder)"
                    ]
            },
              ...
            ]
                \end{codelist}
            \end{iolist}

    \item \module{Concept Categorizer}
    \begin{iolist}
        \item \textit{Input:}
        \textbf{Concept-source sentence:}
        \pydict{
          clear lungs/The lungs are well expanded and clear.,
          stable cardiomediastinal contours/Cardiomediastinal and hilar contours are stable.,
          stable hilar contours/Cardiomediastinal and hilar contours are stable.,
          sternotomy wires/Median sternotomy wires and mediastinal clips are noted.,
          mediastinal clips/Median sternotomy wires and mediastinal clips are noted.,
          no pleural effusion/There is no pleural effusion or pneumothorax.,
          no pneumothorax/There is no pleural effusion or pneumothorax.
        }

        \textbf{Ontology information:}
        \begin{codelist}
            [
              {
                "query": "clear lungs",
                "concept_id": "444859001",
                "matched_surface": "Lungs in normal arrangement",
                "preferred_desc": "Lungs in normal arrangement (finding)",
                "all_descs": [
                  "Lungs in normal arrangement (finding)",
                  "Lungs in normal arrangement"
                ],
                "ancestors": [
                  "Normal lung (finding)",
                  "Normal respiratory system (finding)",
                  "Lung finding (finding)",
                  "Respiratory finding (finding)",
                  "Finding of thoracic region (finding)",
                  "Lower respiratory tract finding (finding)"
                ]
              },
              {
                "query": "stable cardiomediastinal silhouette",
                "concept_id": "233819005",
                "matched_surface": "Stable angina (disorder)",
                "preferred_desc": "Stable angina (disorder)",
                "all_descs": [
                  "Stable angina (disorder)",
                  "Stable angina"
                ],
                "ancestors": [
                  "Chronic ischemic heart disease (disorder)",
                  "Angina (disorder)",
                  "Chronic chest pain (finding)",
                  "Ischemic heart disease (disorder)",
                  "Chronic heart disease (disorder)"
                ]
              },
              {
                "query": "no pneumothorax",
                "concept_id": "49756002",
                "matched_surface": "Perinatal pneumothorax (disorder)",
                "preferred_desc": "Perinatal pneumothorax (disorder)",
                "all_descs": [
                    "Perinatal pneumothorax (disorder)",
                    "Perinatal pneumothorax"
                ],
                "ancestors": [
                    "Pneumothorax (disorder)"
                   "Neonatal respiratory system disorder (disorder)"
                    "Fetal intrapartum and/or early neonatal disorder (disorder)"
                    "Disorder of pleura AND/OR pleural cavity (disorder)"
                    "Neonatal disorder (disorder)"
                    ]
            },
              ...
            ]
                \end{codelist}
        
        \item \proc{Takes each concept and its ontology information(matched terms, description and the ancestors), along with the protocol (ABCDEF) information, as input and categorizes each concept into an ABCDEF category.}
        \item \textit{Output:}
        \pydict{
            clear lungs/B, 
            stable cardiomediastinal contours/C, 
            stable hilar contours/C, 
            sternotomy wires/F, 
            mediastinal clips/F, 
            no pleural effusion/B, 
            no pneumothorax/B
        }
    \end{iolist}

    \item \module{Structured Report Generator}
    \begin{iolist}
        \item \textit{Input:} \\
        \textbf{Categorized concepts:}
        \pydict{
            clear lungs/B, 
            stable cardiomediastinal contours/C, 
            stable hilar contours/C, 
            sternotomy wires/F, 
            mediastinal clips/F, 
            no pleural effusion/B, 
            no pneumothorax/B
        }
        \textbf{Concept-source sentence:}
        \pydict{
          clear lungs/The lungs are well expanded and clear.,
          stable cardiomediastinal contours/Cardiomediastinal and hilar contours are stable.,
          stable hilar contours/Cardiomediastinal and hilar contours are stable.,
          sternotomy wires/Median sternotomy wires and mediastinal clips are noted.,
          mediastinal clips/Median sternotomy wires and mediastinal clips are noted.,
          no pleural effusion/There is no pleural effusion or pneumothorax.,
          no pneumothorax/There is no pleural effusion or pneumothorax.
        }
        \item \proc{Takes the categorized set of concepts, along with the concepts and their corresponding source sentences, and outputs a structured findings section of the report.}
        \item \textit{Output:}
        \pydict{
            A/No findings, 
            B/The lungs are well expanded and clear with no pleural effusion or pneumothorax., 
            C/Cardiomediastinal and hilar contours are stable., 
            D/No findings, 
            E/No findings, 
            F/Median sternotomy wires and mediastinal clips are noted. 
        }
    \end{iolist}
\end{modulelist}

\end{phaselist}



\section{LLM prompts}
\label{app:prompts}

\begin{figure}[t]
\centering
\begin{promptbox}[Planning Prompt with Self-Correction]
You are in \textbf{FULL PIPELINE PLANNING} mode of a Self-Correction PAO agent.

\textbf{USER'S ORIGINAL QUERY:} \{state.user\_query\} \\
\textbf{AVAILABLE TOOLS:} \{tools\_description\}

\textbf{TASK:}
Create a complete pipeline of actions to fully satisfy the user's query. Be prepared to perform self-correction and replanning when prompted by the Observe phase.

You must respond with a JSON array of action steps in the following format:

\noindent
\begin{minipage}{\linewidth}
\begin{verbatim}
[
  {
    "step_number": 1,
    "tool_name": "tool-name",
    "arguments": {
      "argument-name": "value"
    },
    "description": "What this step accomplishes"
  },
  {
    "step_number": 2,
    "tool_name": "another-tool",
    "arguments": {
      "argument-name": "value"
    },
    "description": "What this step accomplishes"
  }
]
\end{verbatim}
\end{minipage}

\textbf{Requirements:}
\begin{enumerate}
    \setlength\itemsep{0em}
    \item Plan \textbf{all} steps required to complete the user's request.
    \item Ensure each step logically follows from the previous one.
    \item Select the appropriate tool for each step.
    \item Provide valid arguments matching the corresponding tool schema.
    \item Include a concise description of the purpose of each step.
    \item Ensure the resulting pipeline fully satisfies the user's request.
    \item If the Observe phase requests a \textbf{REPLAN}, generate an updated pipeline accordingly.
\end{enumerate}

\textbf{IMPORTANT:}
Respond \textbf{ONLY} with the JSON array. Do not include explanations, markdown, or any additional text.
\end{promptbox}
\caption{The system prompt used for the Planning Agent.}
\label{fig:planning_prompt}
\end{figure}


MedPAO-VL follows a PAO loop logic and strategically employs the LLM engine during both the Plan and Observe states of the agent's operational cycle. In the Plan state, the MCP client queries the LLM by providing contextual information including available tools and any previously executed steps, prompting (Figure \ref{fig:planning_prompt}) the engine to determine the next appropriate tool along with its specific parameters. During the Observe state, the system presents (Figure \ref{fig:observation_prompt}) the LLM with both the tool-generated output and the original user query, requesting an evaluation of whether the response adequately addresses the initial request. Based on this assessment, the LLM decides whether to continue iterating through the cycle or terminate the process and return the current output as final. All the individual tool calls of the system uses different prompt to carry out the tasks: Concept Extraction (Figure \ref{fig:extraction_prompt}), Concept categorization (Figure \ref{fig:abcdef_prompt}) and the Structured report generation (Figure \ref{fig:report_generation_prompt}). The Concept Grounded Report Generator generates a fluent, unstructured narrative from the discrete clinical findings and the image. The generation process is designed as a "concept-driven" synthesis, utilizing a finetuned multimodal model. The prompt used to generate this report is depicted in Figure \ref{fig:multimodal_report_prompt}. Furthermore, the same concept grounded report generator is also used self refinement loop to generate the report of missing findings, as indicated by the prompt in Fig~\ref{fig:reprompt_prompt} .


\begin{figure}
\begin{promptbox}[Observation Prompt with Replanning]
\textit{\# Dynamic Context Injection:} \\
\textit{mode\_context = "This was step X of Y in the planned pipeline."}

\vspace{0.5em} \hrule \vspace{0.5em}

You are operating in the \textbf{OBSERVE phase} of a Plan–Act–Observe (PAO) agent.

\textbf{USER QUERY:} \{state.user\_query\}

\textbf{CURRENT PLAN MODE:} \{state.planning\_mode\}

\textbf{ACTION EXECUTED:} \{state.action\_taken\}

\textbf{TOOL OUTPUT:} \{tool\_result\}

\textbf{ALL PREVIOUS RESULTS:} \{state.action\_results if state.action\_results else 'None'\}

\{mode\_context\}

\textbf{TASK}

Analyze the result of the executed tool and determine the correct next action.

You must evaluate:

\begin{enumerate}
\setlength\itemsep{0em}
\item Whether the tool execution was successful.
\item Whether the output contains the expected information.
\item Whether the user's goal has been fully achieved.
\item Whether regeneration or self-correction is required.
\end{enumerate}

\textbf{Special handling rules}

\begin{itemize}
\item If the comparison tool reports missing concepts, instruct the system to regenerate narratives for those missing concepts.
\item If a tool fails or produces invalid output, retry the same step with corrected arguments.
\item If the current plan is no longer valid due to unexpected outputs, request \textbf{REPLAN}.
\item If all required concepts are covered and a structured report is generated, mark the task as complete.
\end{itemize}

Respond strictly in the following format:

\textbf{ANALYSIS:} \\
Your reasoning about the result.

\textbf{GOAL ACHIEVED:} \\
YES or NO.

\textbf{NEXT ACTION:} \\
One of the following:
\begin{itemize}
\item Execute the next planned step.
\item Retry the previous tool with corrected input.
\item Regenerate missing concept narratives.
\item REPLAN if the current plan is invalid.
\item COMPLETE if the goal has been achieved.
\end{itemize}

\end{promptbox}
\caption{Observation prompt used by the PAO agent to analyze tool outputs and decide the next step, including retry and replanning mechanisms.}
\label{fig:observation_prompt}
\end{figure}

\begin{figure}
\resizebox{0.9\textwidth}{!}{
\begin{promptbox}[Concept Extraction Few-Shot Prompt]
\textbf{INSTRUCTION:} Assume you are an expert radiology practitioner. You have to extract the medical concepts from the given report and indicate which sentence each concept came from.

\vspace{0.5em} \hrule \vspace{0.5em}

\textit{\# Example 1 (Negative findings)} \\
Extracted concepts and their corresponding sentence from "No focal consolidation, pleural effusion, or pneumothorax is seen. Heart and mediastinal contours are within normal limits. There is no evidence for pulmonary edema." are \\
\{ \\
\hspace*{1em} "no focal consolidation": "No focal consolidation, pleural effusion, or pneumothorax is seen.", \\
\hspace*{1em} "no pleural effusion": "No focal consolidation, pleural effusion, or pneumothorax is seen.", \\
\hspace*{1em} "no pneumothorax": "No focal consolidation, pleural effusion, or pneumothorax is seen.", \\
\hspace*{1em} "normal heart contours": "Heart and mediastinal contours are within normal limits.", \\
\hspace*{1em} "no pulmonary edema": "There is no evidence for pulmonary edema." \\
\} \#\#

\vspace{0.5em}
\centerline{\textit{[... Examples 2, 3, and 4 omitted for brevity ...]}}
\vspace{0.5em}

\textit{\# Example 5 (Complex pathology)} \\
Extracted concepts and their corresponding sentence from "Cardiac silhouette remains moderately enlarged slightly increased from prior exam. There has been interval increase in central pulmonary vascular engorgement as well as interstitial edema. A focal right lower lung consolidation has increased in severity and is worrisome for pneumonia. [...]" are \\
\{ \\
\hspace*{1em} "enlarged cardiac silhouette": "Cardiac silhouette remains moderately enlarged slightly increased from prior exam.", \\
\hspace*{1em} "pulmonary vascular engorgement": "There has been interval increase in central pulmonary vascular engorgement as well as interstitial edema.", \\
\hspace*{1em} "interstitial edema": "There has been interval increase in central pulmonary vascular engorgement as well as interstitial edema.", \\
\hspace*{1em} "pneumonia": "A focal right lower lung consolidation has increased in severity and is worrisome for pneumonia.", \\
\hspace*{1em} "no pleural effusion": "There is no large pleural effusion or pneumothorax." \\
\} \#\#

\vspace{0.5em} \hrule \vspace{0.5em}

\textbf{TARGET QUERY:} \\
Extracted concepts and their corresponding sentence from \{medical\_final\} are
\end{promptbox}
}
\caption{The few-shot prompt used for extracting clinical concepts and mapping them to their source sentences. The model is conditioned with 5 diverse examples ranging from normal findings to complex pathologies.}
\label{fig:extraction_prompt}
\end{figure}

\begin{figure}
\centering
\begin{promptbox}[ABCDEF Categorization Prompt]
\textbf{SYSTEM ROLE:} You are a radiology expert with the following knowledge:

<ABCDEF Protocol Description>

\vspace{0.5em} \hrule \vspace{0.5em}

\textbf{TASK:} You will be given:
\begin{enumerate}
    \setlength\itemsep{0em}
    \item A concept-source sentence extracted from a radiology report.
    \item A list of medical concepts with their ontology information.
    \item A protocol category.
\end{enumerate}

Your task is to categorize each medical concept into one of the protocol categories: A, B, C, D, E, or F. Use \textbf{both} the concept-source sentence for contextual understanding, and the ontology information (description and ancestors) for semantic understanding.

Ontology information follows the structure:

\noindent
\begin{minipage}{\linewidth}
\begin{verbatim}
{
  "concept": "<medical concept>","description":"<ontology description>",
  "ancestors": ["<ancestor 1>","<ancestor 2>",...]
}
\end{verbatim}
\end{minipage}

Carefully analyze the concept, its ontology description, its ancestors, and the report context before assigning a category.

\vspace{0.5em} \hrule \vspace{0.5em}

\textbf{TARGET QUERY — now categorize all concepts.}

\textbf{Concept-source sentence:} \\
\{concepts\_report\}

\textbf{Ontology data:} \\
\{ontology\_data\}

\textbf{OUTPUT FORMAT:} Return ONLY a valid JSON object in the format:

\noindent
\begin{minipage}{\linewidth}
\begin{verbatim}
{
  "<concept_1>": "<category>",
  "<concept_2>": "<category>"
}
\end{verbatim}
\end{minipage}

Do not provide explanations, reasoning, markdown, or additional text.
\end{promptbox}
\caption{The prompt used to classify extracted concepts into the standard ABCDEF radiological schema. The <ABCDEF Protocol Description> placeholder contains same protocol information as shown in Fig~\ref{fig:report_generation_prompt}}
\label{fig:abcdef_prompt}
\end{figure}


\begin{figure}
\centering
\begin{promptbox}[Structured Report Generation Prompt]
\textbf{SYSTEM ROLE:} You are given categorized medical concepts according to the P-ABCDEF protocol, each mapped to its source sentence from the text report. Chest X-ray review uses A, B, C, D, E, F as a systematic protocol: \textbf{A} airways; \textbf{B} breathing (lungs, pulmonary vessels, pleural spaces); \textbf{C} circulation (cardiomediastinal contour, great vessels); \textbf{D} diaphragm and below; \textbf{E} external (chest wall, soft tissues); \textbf{F} foreign material (devices, foreign bodies).

\vspace{0.3em} \hrule \vspace{0.3em}

\textbf{TASK:} Generate a structured medical report according to the categories above, given the categorized concepts and their source sentences.

\textbf{REQUIREMENTS:}
\begin{enumerate}
    \setlength\itemsep{0em}
    \item \textbf{Style:} Write in a descriptive medical radiologist style.
    \item \textbf{Constraint:} \textit{DO NOT} just aggregate the concepts. \textit{DO NOT} bring in any extra medical findings.
    \item \textbf{Missing Data:} If a category has no concepts, write \texttt{"No Findings"}.
\end{enumerate}

\textbf{OUTPUT FORMAT:}
\noindent
\begin{minipage}{\linewidth}
\begin{verbatim}
{
 "A": "Findings of concepts belonging to A", 
 "B": "...", "C": "...", "D": "...", 
 "E": "...", "F": "..."}
\end{verbatim}
\end{minipage}

\vspace{0.3em} \hrule \vspace{0.3em}

\textbf{TARGET INPUTS:} \\
\textbf{Concept-Source Mapping:} \{concepts\_report\} \\
\textbf{Concept Categories:} \{concepts\}

\textbf{GENERATED REPORT:}
\end{promptbox}
\caption{The prompt used to generate the final structured radiology report. The model is conditioned on the ABCDEF protocol to ensure clinical completeness and is explicitly constrained to avoid hallucinating information not present in the input concepts.}
\label{fig:report_generation_prompt}
\end{figure}


\begin{figure}
\begin{promptbox}[Concept Grounded Report Generation Prompt]
\textbf{LLM ROLE:} Expert radiologist.

\vspace{0.5em} \hrule \vspace{0.5em}

\textbf{USER'S MULTIMODAL INPUT:}

\begin{enumerate}
    \setlength\itemsep{0em}
    \item \textbf{Type:} Image \\
          \textbf{Content:} [A chest X-ray image is provided here]
    \item \textbf{Type:} Text \\
          \textbf{Content:} """You are an expert radiologist analyzing a chest X-ray. Based on the image and observed clinical concepts, generate a comprehensive radiology report. Contain only the findings and impression.
          \textbf{Observed Concepts:}
          \{Present\_concepts, Absent\_concepts\}
          \textbf{Findings:}"""
\end{enumerate}

\vspace{0.5em} \hrule \vspace{0.5em}

\textbf{TASK:} Generate a comprehensive radiology report based on the provided image and clinical concepts.

\textbf{OUTPUT CONSTRAINT:} Report must contain only the findings and impression.
\end{promptbox}
\caption{The multimodal prompt for concept grounded radiology report generation. The model receives a chest X-ray image and a textual prompt containing observed clinical concepts (pre-processed to list present and absent findings), and is tasked with generating a comprehensive report.}
\label{fig:multimodal_report_prompt}
\end{figure}


\begin{figure}
\centering
\begin{promptbox}[Re-prompt Report Generation Prompt]
\textbf{LLM ROLE:} Expert chest radiologist.

\vspace{0.3em} \hrule \vspace{0.3em}

\textbf{CONTEXT:} A previous radiology report missed some findings. Carefully re-check the chest X-ray \textbf{ONLY} for the concepts listed below, and generate the missing part of the report.

\textbf{RULES:}
\begin{enumerate}
    \setlength\itemsep{0em}
    \item Mention \textbf{ONLY} the listed concepts.
    \item If a concept is visible, describe it clearly.
    \item If absent findings are truly absent, explicitly state their absence.
    \item Do not generate a full report.
    \item Do not hallucinate unrelated findings.
    \item Return concise findings only; do not repeat sentences.
    \item Write it professionally, in completely free-style prose --- no numbered or bulleted sections.
    \item \textbf{DO NOT} bold or highlight anything; output must be simple plain text.
\end{enumerate}

\vspace{0.3em} \hrule \vspace{0.3em}

\textbf{TARGET INPUTS:} \\
\textbf{Previously Generated Report:} \{previous\_report\} \\
\textbf{Missing Concepts:} \{missing\_concepts\}

\textbf{FOCUSED FINDINGS:}
\end{promptbox}
\caption{The prompt used in the re-prompting step of the self-correction loop. Given the previously generated report and the set of concepts flagged as missing by the overlap checker, the model is constrained to generate only the missing findings, without repeating or regenerating the full report.}
\label{fig:reprompt_prompt}
\end{figure}
\end{document}